\documentclass[10pt,twocolumn,letterpaper]{article}

\usepackage[pagenumbers]{cvpr} 

\usepackage[utf8]{inputenc} 
\usepackage[T1]{fontenc}    
\usepackage{url}            
\usepackage{booktabs}       
\usepackage{amsfonts}       
\usepackage{amsmath}
\usepackage{nicefrac}       
\usepackage{microtype}      
\usepackage{xcolor}         

\usepackage[utf8]{inputenc} 
\usepackage{CJKutf8} 
\usepackage{graphicx}
\usepackage{xspace}

\usepackage{makecell}
\usepackage{multirow}
\usepackage{tcolorbox}
\definecolor{mycolor_green}{HTML}{E6F8E0}
\usepackage{amssymb}

\definecolor{rred}{RGB}{245, 152, 153}
\definecolor{oorange}{RGB}{253, 205, 154}

\definecolor{cvprblue}{rgb}{0.21,0.49,0.74}
\usepackage[pagebackref,breaklinks,colorlinks,allcolors=cvprblue]{hyperref}

\definecolor{mycolor_blue}{HTML}{00B0F0}
\definecolor{mycolor_purple}{HTML}{FF43EB}
\definecolor{oorange}{RGB}{253, 205, 154}

\usepackage{url}
\usepackage{hyperref}
\hypersetup{
colorlinks,
linkcolor={red},
urlcolor={purple}
}

\def\paperID{5832} 
\def\confName{CVPR}
\def\confYear{2026}

\title{Precise Object and Effect Removal with Adaptive Target-Aware Attention}

\author{Jixin Zhao$^*$ \quad Zhouxia Wang \quad Peiqing Yang \quad Shangchen Zhou$^{*\dag}$ \\
S-Lab, Nanyang Technological University \\
{\tt\small \url{https://zjx0101.github.io/projects/ObjectClear}}
}

\begin{document}

\twocolumn[{%
\renewcommand\twocolumn[1][]{#1}%
\maketitle
\vspace{-10mm}
\begin{center}
    \centering
    \includegraphics[width=\linewidth]{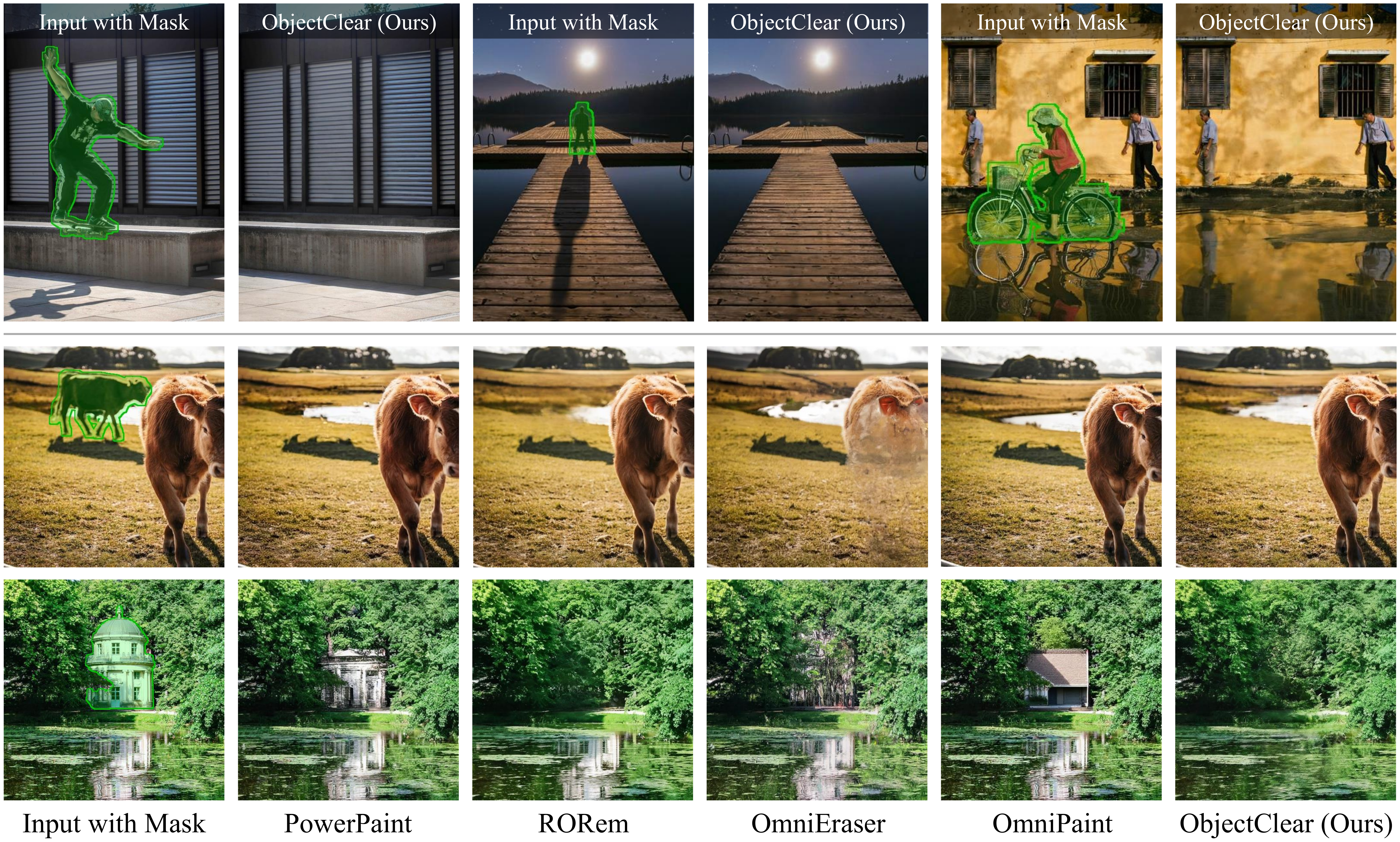}
    \vspace{-5.3mm}
    \captionof{figure}{%
    \textbf{Object Removal Comparison.} Given an object mask, prior methods often leave residual artifacts or hallucinate undesirable content, change background, and usually fail to remove associated effects such as \textit{shadows} and \textit{reflections}. In contrast, our \textit{ObjectClear} precisely eliminates both the object and its associated effects, achieving seamless object removal results even in challenging cases.
    } 
    \vspace{3mm}
    \label{fig:teaser}
\end{center}%
}]

\def\thefootnote{}\footnotetext{\noindent \hspace{-2mm} $^*$Equal contribution \quad $^\dag$Corresponding author
}

\begin{abstract}
Object removal requires eliminating not only the target object but also its associated visual effects such as shadows and reflections. However, diffusion-based inpainting and removal methods often introduce artifacts, hallucinate contents, alter background, and struggle to remove object effects accurately. To address these challenges, we propose \textit{ObjectClear}, a novel framework that decouples foreground removal from background reconstruction via an adaptive target-aware attention mechanism. This design empowers the model to precisely localize and remove both objects and their effects while maintaining high background fidelity. Moreover, the learned attention maps are leveraged for an attention-guided fusion strategy during inference, further enhancing visual consistency. To facilitate the training and evaluation, we construct \textit{OBER}, a large-scale dataset for OBject-Effect Removal, which provides paired images with and without object-effects, along with precise masks for both objects and their effects. The dataset comprises high-quality captured and simulated data, covering diverse objects, effects, and complex multi-object scenes. Extensive experiments demonstrate that \textit{ObjectClear} outperforms prior methods, achieving superior object-effect removal quality and background fidelity, especially in challenging scenarios. 
\end{abstract}    
\vspace{-10mm}
\section{Introduction}
\label{sec:introduction}
Recent advances in generative models~\citep{rombach2022sd, podell2023sdxl,meng2021sdedit}, have demonstrated strong capabilities in image editing and are widely adopted in real-world applications. Among these tasks, object removal has become a key application, enabling users to erase unwanted content from images. However, seamlessly removing both the object and its associated visual effects (\eg, shadows and reflections) while preserving the background remains a challenging problem, as it requires precisely localizing the object together with its effects, restoring the removed regions with background visual consistency, and maintaining the rest of the image unchanged.

With the rapid development of diffusion-based image generation models~\citep{ddpm, podell2023sdxl, flux2024, labs2025flux1kontextflowmatching} and object segmentation techniques~\citep{sam, sam2}, combining diffusion-based image generators with target object masks has become the mainstream paradigm for object removal~\citep{podell2023sdxl, zhuang2024powerpaint, ju2024brushnet, li2025rorem, sun2025attentive, ekin2024clipaway, jia2024designedit, chen2024freecompose}.
However, as shown in Figs.~\ref{fig:teaser} and~\ref{fig:sota_wild}, existing methods often fail to remove the visual effects of target objects, generate undesired objects in the removed areas, or unintentionally modify the background (see Fig.~\ref{fig:bg_diff}). 
These limitations mainly stem from the lack of explicit modeling of the correlation between the target object and its associated effects, as well as the absence of effective constraints to guide the generative model’s attention toward the removal regions.

To address these challenges, we propose ObjectClear, a novel framework designed to achieve precise and complete removal of both objects and their visual effects while preserving background fidelity. To enhance the modeling of object–effect regions, we introduce an Adaptive Target-Aware Attention (ATA) mechanism that learns to localize objects and their associated effects under the supervision of object–effect masks.
By integrating text–image embeddings, ATA effectively decouples foreground removal from background reconstruction, enabling the model to remove targets accurately and suppress undesired object generation within the removed regions.
During inference, we further introduce an Attention-Guided Fusion (AGF) module that leverages ATA-predicted object–effect attention maps to perform adaptive input–output fusion, preserving fine-grained background details including color and texture. In addition, we introduce a Spatially-Varying Denoising Strength (SVDS) strategy to address incomplete removal and color inconsistency caused by uniform denoising strength. Together, these designs empower ObjectClear to deliver precise and high-fidelity effect-object removal.

To train ObjectClear and fully realize its potential, a dedicated dataset containing paired images with and without objects, along with precise masks for both objects and their effects, is essential. Unfortunately, to the best of our knowledge, no such large-scale training dataset is publicly available.
Current object removal datasets can be categorized into \textit{simulated} and \textit{camera-captured} data. (1)~\textit{Simulated data.} These datasets are often constructed by copy-pasting objects~\citep{jiang2025smarteraser, li2024magiceraser} or using pretrained inpainting models to generate pseudo ground truth~\citep{tudosiu2024mulan}. While this allows low-cost generation of large-scale data, such datasets typically lack object effects such as shadows and reflections, causing models trained on them to struggle with effect removal. (2)~\textit{Camera-captured data.} 
Some works~\citep{Sagong2022rord, wei2025omnieraser} extract paired frames from fixed-viewpoint videos, but this restricts the foreground to moving objects only and makes it difficult to maintain background consistency between paired samples.
Others~\citep{winter2024objectdrop, yu2025omnipaint, yang2024layerdecomp} capture image pairs before and after object removal, but these datasets remain small in scale due to their high collection cost and are not publicly released.

To overcome these limitations, we present \textit{OBER}, a hybrid dataset for \textbf{OB}ject-\textbf{E}ffect \textbf{R}emoval, which combines both \textit{camera-captured} and \textit{simulated} data with diverse foreground objects, background scenes, and object effects (\eg, indoor/outdoor shadows and reflections). \textit{For the camera-captured data}, we annotate precise object and object-effect masks, which serve as critical supervision during training. \textit{For the simulated data}, we leverage these masks to compute accurate alpha maps, enabling realistic alpha blending of RGBA object layers with effects and high-quality backgrounds. We further extend the simulation pipeline to include multi-object scenarios, enhancing model robustness in challenging cases involving occlusion and object interactions, as shown in the last two rows of Fig.~\ref{fig:sota_wild}. 
By combining the realism of captured data with the scalability of simulation, OBER offers a \textit{high-quality}, \textit{large-scale}, and \textit{diverse} dataset with a total of 12,715 training samples. In addition, we introduce two new benchmarks: \textit{OBER-Test} and \textit{OBER-Wild}, to support future research. 
 
Our contributions are summarized as follows: 
(1) We introduce a novel framework, \textit{ObjectClear}, which incorporates an Adaptive Target-Aware Attention (ATA) mechanism that adaptively focuses on foreground removal regions, together with an Attention-Guided Fusion (AGF) and a Spatially-Varying Denoising Strength (SVDS) strategy, thereby improving object removal quality and background fidelity.
(2) We propose \textit{OBER}, a high-quality and large-scale hybrid dataset for object-effect removal, featuring diverse objects, fine-grained annotations of object-effect masks, and complex scenarios across both simulated and camera-captured settings.
(3) ObjectClear achieves superior performance on both synthetic and real-world benchmarks, demonstrating clear advantages in quantitative results and visual quality.
\section{Related Work}
\label{sec:related_work}

\begin{figure*}[t!]
\begin{center}
    \vspace{-3mm}
    \includegraphics[width=\linewidth]{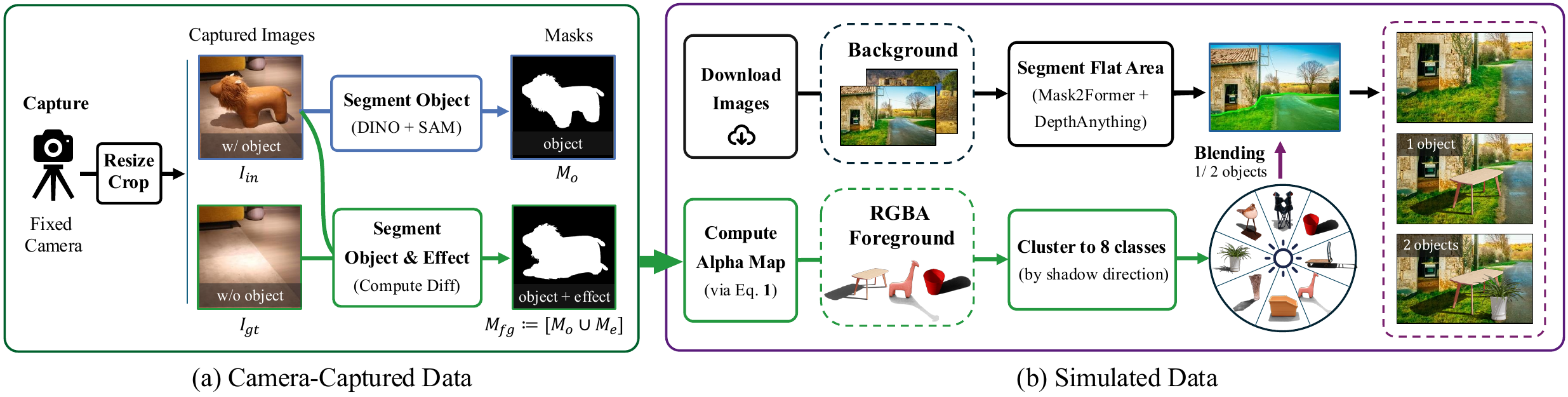}
    \vspace{-5mm}
    \caption{    
    \textbf{Dataset Construction Pipeline of \textit{OBER}.}
    The dataset combines both \textit{camera-captured} and \textit{simulated} data, featuring diverse foreground objects and background scenes. It provides rich annotations, including object masks, object-effect masks, transparent RGBA object layers, and complex multi-object scenarios for training and evaluation.
    }
    \vspace{-6mm}
\label{fig:data}
\end{center}
\end{figure*}

\noindent{\bf Image Inpainting.}
Image inpainting is a long-standing visual editing task that aims to seamlessly reconstruct pixels within a given mask. 
Early approaches predominantly adopt generative adversarial networks (GANs)~\citep{goodfellow2014generative}, but often suffer from limited realism and diversity~\citep{liu2020rethinking, pathak2016context, ren2019structureflow, zeng2019learning}. 
With the rapid advances in diffusion models~\citep{rombach2022sd, song2020score, ho2020denoising, podell2023sdxl}, many methods~\citep{avrahami2022blended, lugmayr2022repaint, meng2021sdedit, zhuang2024powerpaint, ju2024brushnet, yang2023pgdiff, wang2025towards} have begun leveraging their strong generative priors to synthesize high-fidelity content, achieving state-of-the-art results in image inpainting. 
In this work, we adapt the SDXL-Inpainting model~\citep{podell2023sdxl} for photorealistic completion. 
However, despite their strong generative capabilities, existing inpainting models often lack awareness of object effects (\textit{e.g.}, shadows and reflections), leading to incomplete or inconsistent object removal results.

\noindent{\bf Object Removal.}
Object removal is a specialized branch of image inpainting that often requires explicit consideration of object effects (\eg, shadows and reflections) to achieve complete removal, a problem increasingly dominated by diffusion-based methods~\citep{winter2024objectdrop, jiang2025smarteraser, liu2025erasediff, yu2025omnipaint, li2025rorem, chen2024unireal, zhuang2024powerpaint, ekin2024clipaway, chen2024freecompose, yang2024layerdecomp, jia2024designedit, sun2025attentive, wei2025omnieraser, zhu2025georemover}, and recently extended to the video domain~\citep{zhou2023propainter, Lee2025generativeomnimatte, miao2025rose, samuel2025omnimattezero}.
A common strategy is to curate triplet datasets. ObjectDrop~\citep{winter2024objectdrop} builds a real-world dataset by capturing the same scene before and after object removal, but its limited scale and non-public availability hinder broader adoption.
To improve data scalability, methods like SmartEraser~\citep{jiang2025smarteraser} and EraseDiffusion~\citep{liu2025erasediff} rely on synthetic datasets generated by segmentation or matting methods to extract foreground objects, but these datasets typically lack annotations of object effects, limiting the ability to remove shadows or reflections. OmniPaint~\citep{yu2025omnipaint} further expands its dataset by automatically annotating unlabeled images using a model trained on small-scale real data, while RORem~\citep{li2025rorem} also incorporates human annotators to ensure annotation quality.
In parallel, works like RORD~\citep{Sagong2022rord} and OmniEraser~\citep{wei2025omnieraser} scale data generation by mining realistic video frames with fixed viewpoints, selectively pairing frames w/ and w/o target objects while preserving natural object effects.
GeoRemover~\citep{zhu2025georemover} leverages depth geometry to achieve precise object removal.
To eliminate reliance on curated datasets altogether, other methods opt for test-time optimization~\citep{chen2024freecompose, sun2025attentive, jia2024designedit} or providing plug-and-play solutions~\citep{ekin2024clipaway}.
However, these models rely on implicitly learning object effects without explicitly modeling effect maps, which makes it difficult to maintain background consistency.

%
\begin{figure*}[!t]
\begin{center}
    \vspace{-3mm}
    \includegraphics[width=0.9\linewidth]{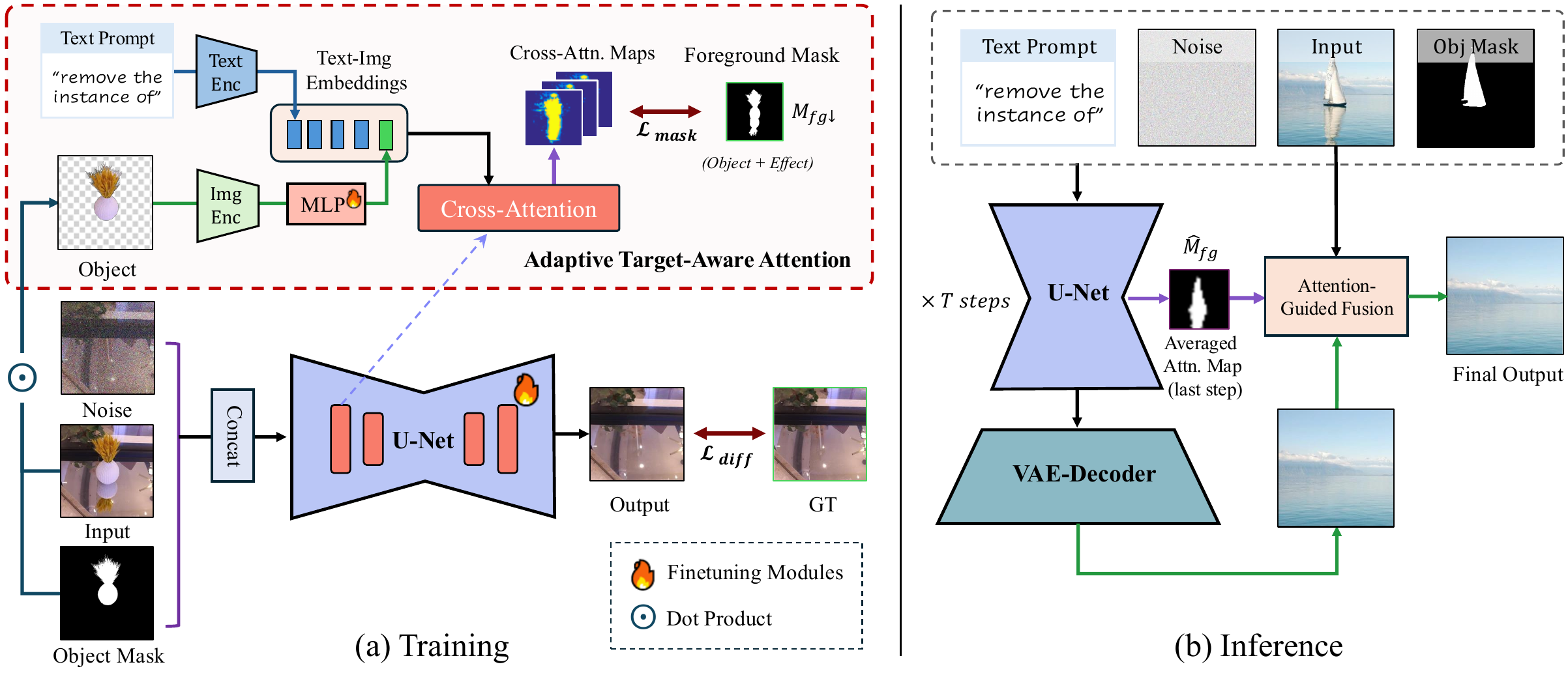}
    \vspace{-1mm}
    \caption{    
    \textbf{The Framework of \textit{ObjectClear}.} Given an input image and a target object mask, \textit{ObjectClear} employs an Adaptive Target-Aware Attention mechanism to guide the model toward foreground removal regions by learning attention masks. The predicted mask further enables an Attention-Guided Fusion strategy during inference, which substantially preserves background details.
    }
    \vspace{-6mm}
\label{fig:network}
\end{center}
\end{figure*}

\section{Methodology}
\label{sec:method}

To achieve accurate removal of target objects and their associated effects, we propose a comprehensive framework with a data curation pipeline (OBER Dataset, Sec.~\ref{sec:data}) and a dedicated object removal model (ObjectClear, Sec.~\ref{sec:objectclear}).

\vspace{-0.5mm}
\subsection{OBER Dataset}
\label{sec:data}
\vspace{-0.5mm}
To learn an object–effect removal model, a high-quality dataset with paired images and precise masks for both objects and their effects is required. However, to our knowledge, no such large-scale dataset is publicly available.
To this end, we introduce the \textit{OBER} dataset (\textbf{OB}ject-\textbf{E}ffect \textbf{R}emoval), designed to balance data realism and scalability. 
As shown in Fig.~\ref{fig:data}, OBER is a \textit{hybrid} dataset consisting of two parts: (1) a small set (2,878) of camera-captured images adhering to physical realism, and (2) a larger set (10,000) of simulated images generated by compositing foreground objects, extracted from (1), onto diverse background scenes.

\noindent{\bf Camera-Captured Data.} 
\textit{\textbf{(1)~Capture Paired Images.}}
Following the approach of ObjectDrop~\citep{winter2024objectdrop}, we use fixed cameras to construct a counterfactual dataset by capturing each scene before and after the removal of a single object, while keeping all the other factors unchanged. For each pair, the image with the object is used as the input $I_{in}$, and the image without the object serves as the ground truth $I_{gt}$.
In total, we collected {2,878} such counterfactual pairs with 2,715 for training and 163 for testing. Our dataset encompasses a wide variety of everyday objects commonly found in indoor (e.g., benches, cups, dolls) and outdoor scenes (e.g., pedestrians, vehicles). These image pairs also preserve visual effects such as shadows and reflections. To reduce potential pixel-level misalignment, we downsample and crop all images to a fixed training resolution (i.e., 512 $\times$ 512).
\textit{\textbf{(2)~Segment Object Mask $M_o$.}}
To obtain the object mask $M_o$ as network input alongside the input image $I_{in}$, we apply off-the-shelf detection and segmentation models, such as DINO~\citep{dino} and SAM~\citep{sam} to $I_{in}$.
\textit{\textbf{(3)~Segment Object-Effect Mask $M_o \cup M_e$.}}
Different from previous methods, we propose to explicitly model the object effects. Therefore, we introduce an object-effect mask $M_{fg}$ that covers both the object $M_o$ and its associated visual effects $M_e$ in our dataset. Unlike the coarse object-effect masks provided in the RORD dataset~\citep{Sagong2022rord}, we construct $M_{fg}$ in an efficient and accurate way by computing the pixel-wise difference between $I_{in}$ and $I_{gt}$. Pixels with differences above a predefined threshold are regarded as part of the object-effect mask.
This object-effect mask provides crucial supervision during training, allowing the network to adaptively learn to focus on the target removal regions, including both the object itself and its associated effects.

\noindent{\bf Simulated Data.}
With the high-quality real data collected, we further scale up the dataset with a carefully designed simulation pipeline.
\textit{\textbf{(1) Collect Background Images.}}
We begin by downloading high-quality background images from the Internet, which offer a diverse range of background scenes. To select backgrounds with flat regions suitable for object placement, we first apply Mask2Former~\citep{mask2former} to segment flat areas corresponding to semantic classes such as ``\textit{road, sidewalk, grass, floor}'', as illustrated in Fig.~\ref{fig:data}(b). We then refine the selection by computing the gradient of the depth map generated by Depth Anything V2~\citep{depthanythingv2}, filtering out regions with significant depth variation to ensure that the inserted objects are placed on visually flat surfaces.
\textit{\textbf{(2) Collect Foreground Objects with Effects.}}
Based on the camera-captured paired data $(I_{in}, I_{gt})$, along with the object mask $M_o$ and effect mask $M_e$, we compute the alpha map of the foreground object with effects $I_{oe}$ using Eq.~\ref{eq:alpha}, where $\varepsilon$ is a small constant added to avoid division by zero. For subsequent compositing, we manually categorize the foreground objects into eight groups based on their shadow direction (Fig.~\ref{fig:data}(b)).
\begin{equation}
\alpha(p) =
\begin{cases}
0, &  \text{if } p \in \text{$\overline{M_o} \cap \overline{M_e}$} \\
1, &  \text{if } p \in \text{$M_o$} \\
(I_{\text{gt}} - I_{\text{in}}) / (I_{\text{gt}} + \varepsilon), &  \text{if } p \in \text{$M_e$} \\
\end{cases},
\label{eq:alpha}
\end{equation}
where $M_o$ and $M_e$ are object and effect regions, respectively.

\noindent \textit{\textbf{(3) Blend Objects with Backgrounds.}}
We randomly sample a background image $I_{bg}$ and a foreground object image with effects $I_{oe}$, and synthesize a composite image using alpha blending: $I_{comp} = (1 - \alpha) \cdot I_{bg} + \alpha \cdot I_{oe}$.
Beyond single-object cases, we also synthesize \textit{multi-object} data by compositing multiple foreground objects with the same lighting directions, thereby covering scenarios involving object occlusions, as illustrated in Fig.~\ref{fig:data}(b).
These designs ensure physically plausible placement and consistent lighting, reducing the domain gap between simulated and real data. Synthetic data also overcomes challenges in real-world collection, such as limited scale and complex cases (e.g., multi-object occlusions, reflections), while its controllability expands coverage and diversity.
In total, we generate {10,000} composite images, significantly enriching the diversity and scalability of our dataset. In particular, the simulation of multi-object compositions leads to notable improvements in object removal robustness and background preservation, as discussed and compared in Table~\ref{tab:ablation}.

\subsection{ObjectClear}
\label{sec:objectclear}

Our ObjectClear is built upon SDXL-Inpainting~\citep{podell2023sdxl}. While SDXL-Inpainting uses a noise map $z_t$, a masked image $I_{m}$, a corresponding mask $M_o$, and a text prompt $c$ as inputs, we feed the original image $I_{in}$ instead of $I_m$ into the model, and our inputs are expressed as a tuple $<z_{t}, I_{in}, M_{o}, c>$. This design encourages the model to better attend to the effect of the target object by leveraging its visual features. Moreover, it facilitates more effective utilization of background information behind the object when transparent objects are to be removed, such as glass cups.
To achieve precise and complete object removal while enhancing background preservation, we introduce Adaptive Target-Aware Attention (ATA), Attention-Guided Fusion (AGF), and Spatially-Varying Denoising Strength (SVDS) strategy, which explicitly attend to object-effect regions.

\noindent{\bf Adaptive Target-Aware Attention.}
To enable the model to better attend to both the target object region and its associated effect regions, we integrate text and object image as an object prompt for the cross-attention layers in the base model. Specifically, the text prompt is expressed as ``\textit{remove the instance of}'', and the visual object is obtained by applying a dot product between the input image $I_{in}$ and $M_o$, as illustrated in Fig.~\ref{fig:network}(a). These two modalities are then encoded into text embeddings and visual embeddings using the CLIP~\citep{clip} text and vision encoders, respectively. To unify their representation spaces, the visual embeddings are further projected into the same dimensional space as the text embeddings using a lightweight trainable MLP composed of two linear layers. The resulting text embeddings and projected object embeddings are then stacked and used as guidance in the cross-attention blocks of the base model. 
To encourage the model to focus more accurately on the object and its effect regions, we introduce a mask loss $\mathcal{L}_{mask}$. Concretely, we extract the cross-attention maps corresponding to the visual embedding tokens and denote them as $\mathbf{A}$, which we supervise with the annotated foreground object-effect masks $M_{fg}$ from our OBER dataset. 
$\mathcal{L}_{\text{mask}}$ is designed to minimize the attention values in the background regions while maximizing those in foreground. This objective can be formulated as:
\begin{equation}
\mathcal{L}_{mask} = \mathrm{mean}({\mathbf{A}}[1-M_{fg}]) - \mathrm{mean}({\mathbf{A}}[M_{fg}])\text{,}
\label{eq:oba_loss}
\end{equation}
where $[\cdot]$ denotes indexing operator, and $\mathbf{A}[M_{fg}]$ refers to the attention values within the foreground regions specified by the mask $M_{fg}$.

\noindent{\bf Attention-Guided Fusion.}
Interestingly, we observe that applying Adaptive Target-Aware Attention not only improves ObjectClear’s precision in object removal, but also produces attention maps that accurately capture both the object and its effects. Based on this observation, we propose an Attention-Guided Fusion strategy that leverages the predicted attention maps during inference to seamlessly blend the generated result with the original image. Specifically, we extract the first-layer cross-attention map corresponding to the object embedding during inference. This attention map is then upsampled to match the resolution of the original image, forming a soft estimate of the object-effect region. To avoid edge artifacts during blending, we apply a Gaussian blur to the upsampled attention map, producing a soft-edged object-effect mask. This mask is then used in an alpha-blending operation to fuse the generated image with the original input.
This strategy significantly reduces undesired background changes introduced by the diffusion denoising process and VAE reconstruction, achieving high-fidelity object removal. Unlike BrushNet~\citep{ju2024brushnet}, which relies on user-provided object masks that usually exclude effect regions, our approach leverages precise object-effect masks adaptively generated by Adaptive Target-Aware Attention.

\noindent{\bf Spatially-Varying Denoising Strength.}
In diffusion-based image editing, the initial latent for denoising is typically obtained by adding a certain amount of noise to the input image latent. The denoising strength (DS) $\mathrm{DS} \in [0,1]$ controls the noise level: a larger value injects more noise, thereby pushing the initial latent $z_t$ closer to the pure-noise prior. 
Specifically, when $\mathrm{DS}=1.0$, the diffusion process starts entirely from noise and discards all information from the inputs, achieving complete object removal but often introducing noticeable global color shifts.
In contrast, using a slightly lower denoising strength (\eg, $\mathrm{DS}=0.99$) preserves color consistency but risks incomplete object removal or hallucinated unwanted objects. Motivated by these observations, we propose Spatially-Varying Denoising Strength (SVDS), which applies $\mathrm{DS}=1.0$ within the masked object region and $\mathrm{DS}=0.99$ in the unmasked background, via re-injecting the background during inference. This strategy achieves complete object-effect removal while maintaining background color consistency and preventing edge artifacts during Attention-Guided Fusion (see ablation in the supp.).

\section{Experiments}
\vspace{-1mm}
\label{sec:experiment}

\begin{table*}[t]
  \vspace{-3mm}
  \centering
  \caption{\textbf{Quantitative Comparisons with State-of-the-Art Methods}. The best and second performances are marked in \colorbox{rred}{\underline{red}} and \colorbox{oorange}{{orange}}, respectively. Although ObjectClear takes only the object mask as input, it outperforms prior methods on most metrics, even when some methods are provided with masks that cover both the object and its associated effect regions.
  }
  \vspace{-2mm}
    \renewcommand{\arraystretch}{1.14}
    \renewcommand{\tabcolsep}{2.0mm}
  \resizebox{\linewidth}{!}{
    \begin{tabular}{@{}c|l|cccc|cccc|c@{}}
    \toprule
    Mask & Datasets 
    & \multicolumn{4}{c|}{RORD-Val} 
    & \multicolumn{4}{c|}{OBER-Test} 
    & OBER-Wild \\
    \cmidrule(lr){3-6} \cmidrule(lr){7-10} \cmidrule(lr){11-11}
    Types & Metrics 
    & PSNR$~\uparrow$ & PSNR-BG$~\uparrow$ & LPIPS$~\downarrow$ & CLIP$~\downarrow$ 
    & PSNR$~\uparrow$ & PSNR-BG$~\uparrow$ & LPIPS$~\downarrow$ & CLIP$~\downarrow$ 
    & ReMOVE$^{\dagger}$$\uparrow$ \\
    \midrule
    \multirow{9}*{\rotatebox{90}{Object-Effect Mask}} & SDXL-INP~\citep{podell2023sdxl} 
    & 19.39 & 22.98 & 0.2432 & 0.1024
    & 24.07 & 26.93 & 0.1296 & 0.0579
    & 0.6983 \\
    ~ & PowerPaint~\citep{zhuang2024powerpaint} 
    & 19.87 & 21.98 & 0.2303 & 0.0776
    & 26.20 & 27.41 & 0.1243 & 0.0409
    & 0.8044 \\
    ~ & BrushNet~\citep{ju2024brushnet} 
    & 16.82 & 18.87 & 0.3434 & 0.1692
    & 20.96 & 23.69 & 0.2052 & 0.1180
    & 0.5358 \\
    ~ & DesignEdit~\citep{jia2024designedit} 
    & 20.69 & 22.49 & 0.2946 & 0.1200
    & 26.59 & 27.63 & 0.1777 & 0.0629
    &  0.8215 \\
    ~ & CLIPAway~\citep{ekin2024clipaway} 
    & 18.87 & 20.78 & 0.3328 & 0.0969
    & 25.38 & 26.28 & 0.1039 & 0.0349
    & 0.7705 \\
    ~ & FreeCompose~\citep{chen2024freecompose} 
    & 19.67 & 21.72 & 0.3316 & 0.1110
    & 23.39 & 25.15 & 0.1305 & 0.0629
    & 0.7555 \\
    ~ & Attentive Eraser~\citep{sun2025attentive} 
    & 20.33 & 21.98 & 0.2545 & 0.1015
    & 27.42 & 29.27 & 0.1114 & 0.0249
    & 0.7940 \\
    ~ & RORem~\citep{li2025rorem} 
    & 21.61 & 23.11 & 0.3224 & 0.0767
    & 27.23 & 27.95 & 0.1042 & 0.0234
    & 0.8164 \\
    \midrule
    \midrule
    \multirow{11}*{\rotatebox{90}{Object Mask}} & SDXL-INP~\citep{podell2023sdxl} 
    & 20.23 & 24.83 & 0.2042 & 0.0868
    & 22.42 & 25.77 & 0.1428 & 0.0771
    & 0.6971 \\
    ~ & PowerPaint~\citep{zhuang2024powerpaint}
    & 21.46 & 24.62 & 0.1801 & 0.0648
    & 22.76 & 24.67 & 0.1544 & 0.0729
    & 0.7699 \\
    ~ & BrushNet~\citep{ju2024brushnet} 
    & 18.06 & 23.44 & 0.2757 & 0.1821
    & 21.19 & 24.38 & 0.1822 & 0.1123
    & 0.6341 \\
    ~ & DesignEdit~\citep{jia2024designedit} 
    & 22.09 & 24.26 & 0.2501 & 0.1021
    & 24.63 & 25.48 & 0.1870 & 0.0788
    & 0.8163 \\
    ~ & CLIPAway~\citep{ekin2024clipaway} 
    & 20.58 & 23.21 & 0.2770 & 0.0785
    & 22.32 & 24.05 & 0.1357 & 0.0765
    & 0.7705 \\
    ~ & FreeCompose~\citep{chen2024freecompose} 
    & 20.39 & 22.91 & 0.3015 & 0.0897
    & 22.77 & 24.46 & 0.1393 & 0.0690
    & 0.7451 \\
    ~ & Attentive Eraser~\citep{sun2025attentive} 
    & 22.17 & 24.59 & 0.1883 & 0.0643
    & 25.70 & 27.08 & 0.1201 & 0.0437
    & 0.8080 \\
    ~ & RORem~\citep{li2025rorem} 
    & 22.49 & 24.10 & 0.2943 & 0.0634
    & 24.51 & 25.28 & 0.1288 & 0.0460
    & 0.8121 \\
    \cline{2-11}
    ~ & OmniEraser~\citep{wei2025omnieraser} 
    & 21.79 & 22.98 & 0.2195 & 0.0542
    & 24.44 & 24.87 & 0.1783 & \colorbox{oorange}{0.0142}
    & 0.7655 \\
    ~ & GeoRemover~\citep{zhu2025georemover} 
    & \colorbox{oorange}{24.38} & \colorbox{oorange}{26.47} & 0.1924 & 0.0575
    & 24.92 & 25.33 & 0.1976 & 0.0670
    & 0.7938 \\
    ~ & OmniPaint~\citep{yu2025omnipaint} 
    & 22.75 & 24.66 & \colorbox{oorange}{0.1178} & \colorbox{oorange}{0.0354}
    & \colorbox{oorange}{29.06} & \colorbox{oorange}{30.04} & \colorbox{oorange}{0.0521} & 0.0152
    & \colorbox{rred}{\underline{0.8656}} \\
    ~ & \textbf{ObjectClear (Ours)} 
    & \colorbox{rred}{\underline{26.24}} & \colorbox{rred}{\underline{29.78}} & \colorbox{rred}{\underline{0.1157}} & \colorbox{rred}{\underline{0.0299}}
    & \colorbox{rred}{\underline{33.04}} & \colorbox{rred}{\underline{35.62}} & \colorbox{rred}{\underline{0.0342}} & \colorbox{rred}{\underline{0.0103}}
    & \colorbox{oorange}{0.8470} \\
    \bottomrule
  \end{tabular}
  }
  \vspace{-1mm}
  \label{tab:sota}
\end{table*}

\begin{figure*}[!t]
    \centering
    \includegraphics[width=\linewidth]{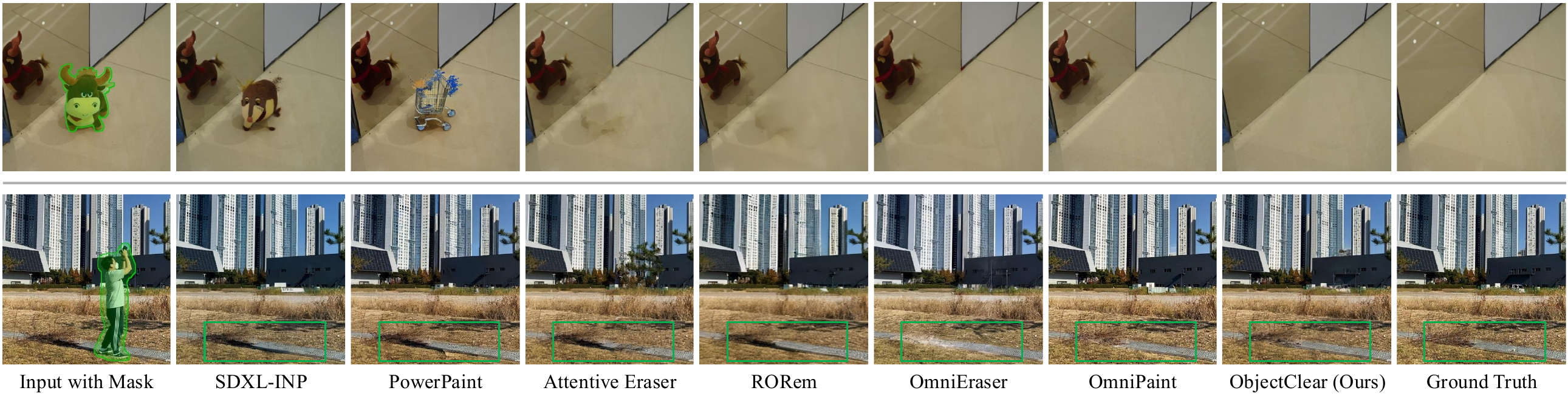}
    \vspace{-5.5mm}
    \caption{\textbf{Object Removal on OBER-Test and RORD-Val.} Our ObjectClear effectively removes both the masked objects and their associated effects, including shadows and mirror reflections.}
    \label{fig:sota_gt}
    \vspace{-3.5mm}
\end{figure*}

\noindent{\bf Implementation.}
Our network is built on SDXL-Inpainting~\citep{podell2023sdxl} and fine-tuned with our OBER dataset with an input resolution of $512 \times 512$. Training is conducted with a total batch size of $32$ on 8 A100 GPUs for 100k iterations and a learning rate of 1e-5. All experiment results are attained with a guidance scale of $1.0$ and $20$ denoising steps. 

\begin{figure*}[t]
    \centering
    \vspace{-3mm}
    \includegraphics[width=\linewidth]{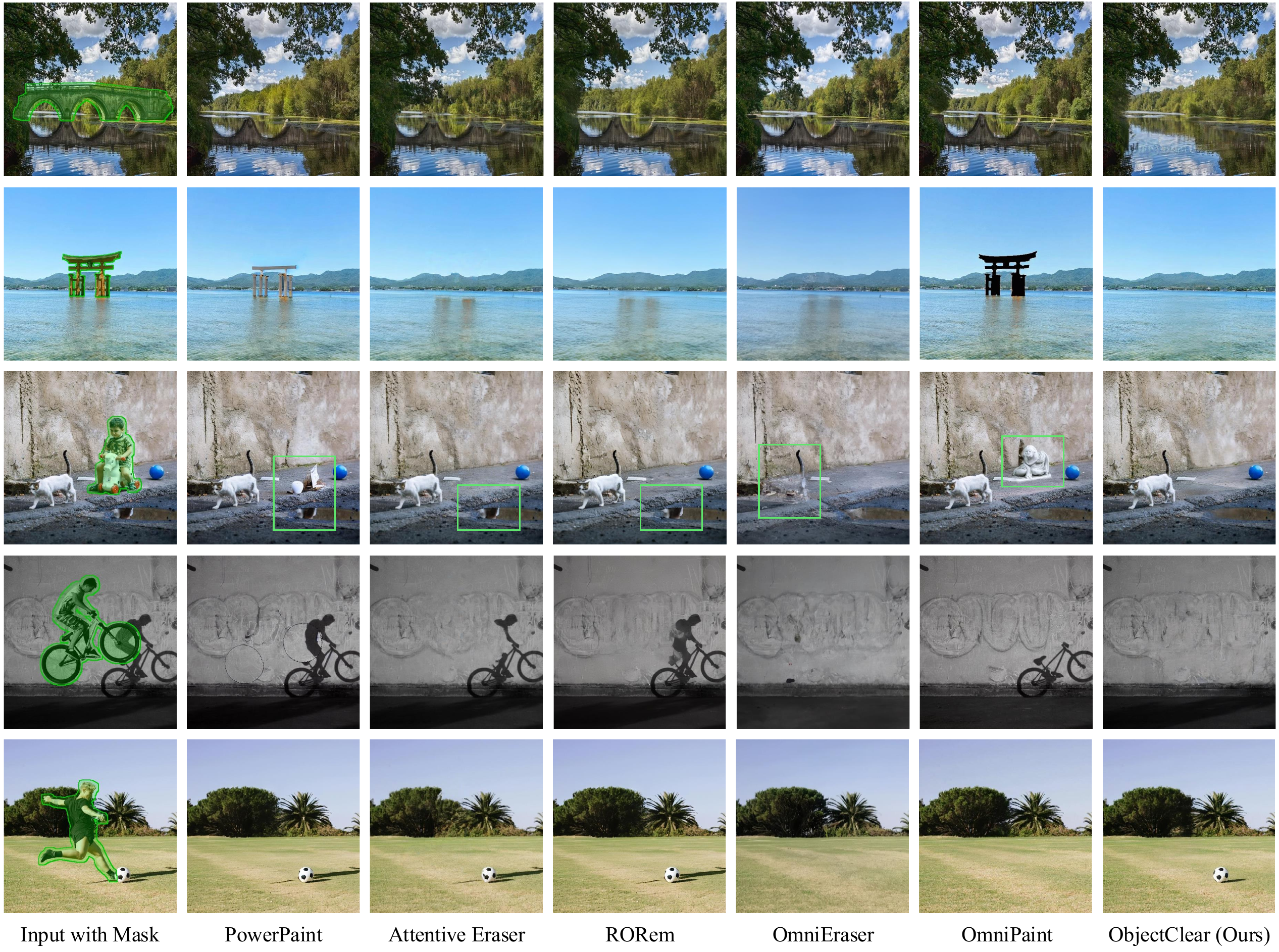}
    \vspace{-5mm}
    \caption{\textbf{Object Removal on OBER-Wild.} 
    Prior methods may not completely remove object effects and can also introduce residual artifacts or unwanted content. In contrast, ObjectClear achieves precise object–effect removal while preserving high background fidelity.}
    \label{fig:sota_wild}
    \vspace{-3mm}
\end{figure*}

\noindent{\bf Evaluation Data.}
We evaluate ObjectClear and prior methods on three test sets:
(1)~\textit{RORD-Val.} RORD~\citep{Sagong2022rord} is a widely used object removal dataset. Each sample contains an image pair (w/ and w/o object) and a coarse mask covering both object and its effect. To avoid duplicated scenarios, we randomly select one image per scene, resulting in a RORD-Val with \textit{343} samples. Since some object removal methods, including OmniEraser~\cite{wei2025omnieraser}, OmniPaint~\cite{yu2025omnipaint} and ours, require only the object mask as input, we manually refine the object masks to ensure fair evaluation.
(2)~\textit{OBER-Test.} As described in Sec.~\ref{sec:data}, we split our OBER dataset into training and test sets. The test set contains 163 samples, each providing image pairs and precise masks for both object and effect.
(3)~\textit{OBER-Wild.} 
To assess generalization in real-world scenarios, we collect 302 in-the-wild images featuring objects with shadows or reflections.
Object masks are annotated using DINO~\citep{dino} and SAM~\citep{sam}, while effect masks are manually labeled.
Ground-truth removal images are not available for this set.

\noindent{\bf Evaluation Metrics.}
For RORD-Val and OBER-Test, where ground truth is available, we evaluate performance using fidelity metrics: PSNR and PSNR-BG (computed only on background regions), as well as two perceptual metrics: LPIPS~\citep{lpips} and CLIP~\citep{clip} (feature distance). For OBER-Wild, which lacks ground truth, we evaluate results using ReMOVE$^{\dagger}$, a modified version of ReMOVE~\citep{remove} designed to better assess visual harmony. Specifically, ReMOVE$^{\dagger}$ measures the consistency between the output’s removal region and the input background, instead of comparing against the output background as in the original ReMOVE.

\vspace{-2mm}
\subsection{Comparisons with State-of-the-Art Methods}
\vspace{-2mm}
We compare ObjectClear with state-of-the-art methods, including: 1) \textit{image inpainting methods} of SDXL-INP~\citep{podell2023sdxl}, PowerPaint~\citep{zhuang2024powerpaint}, and BrushNet~\citep{ju2024brushnet}; 2) \textit{object removal methods} of CLIP-Away~\citep{ekin2024clipaway}, DesignEdit~\citep{jia2024designedit}, RORem~\citep{li2025rorem}, FreeCompose~\citep{chen2024freecompose}, Attentive Eraser~\citep{sun2025attentive}, OmniEraser~\citep{wei2025omnieraser}, GeoRemover~\citep{zhu2025georemover}, and OmniPaint~\citep{yu2025omnipaint}.

\noindent \textbf{Quantitative Evaluation.} Unlike our ObjectClear, which explicitly handles both the target object and its associated effects, most baseline methods operate only within the masked regions, overlooking surrounding areas that are visually affected by the object but not explicitly included in the mask. To ensure a fair and comprehensive comparison, we test all methods under two mask conditions: (1) \textit{Object Mask}, and (2) \textit{Object-Effect Mask},
as reported in Table~\ref{tab:sota}.
ObjectClear achieves state-of-the-art performance on nearly all test sets and metrics. Notably, even when using only the object mask, ObjectClear surpasses methods that rely on both object and effect masks. In particular, it achieves a significant advantage in the PSNR-BG metric, highlighting its superior ability to preserve background consistency with the input.

\begin{table*}[!t]
    \centering
    \vspace{-3mm}
    \caption{\textbf{Quantitative Results of Ablation Studies.} Based on our camera-captured data (CC~Data), object removal performance improves progressively by introducing the Adaptive Target-Aware Attention (ATA), simulated training data (Sim.~Data), Attention-Guided Fusion (AGF), and Spatially-Varying Denoising Strength (SVDS) strategy. 
    }
    \label{tab:ablation}
    \vspace{-1mm}
    \renewcommand{\arraystretch}{1.0}
    \renewcommand{\tabcolsep}{4.0mm}
    \resizebox{\linewidth}{!}{
    \begin{tabular}{c|ccccc|cccc}
        \toprule
        Exp. & CC Data & ATA & Sim.~Data & AGF & DS & PSNR$~\uparrow$ & PSNR-BG$~\uparrow$ & LPIPS$~\downarrow$ & CLIP$~\downarrow$ \\
        \midrule
        (a) & \checkmark & & & & 0.99 & 27.29 & 27.96 & 0.0910 & 0.0247 \\
        (b) & \checkmark & \checkmark & & & 0.99 & 27.56 & 28.37 & 0.0845 & 0.0217 \\
        (c) & \checkmark & \checkmark & \checkmark & & 0.99 & 28.04 & 28.80 & 0.0805 & 0.0196 \\
        (d) & \checkmark & \checkmark & \checkmark & \checkmark & 0.99  & {32.77} & {35.50} & {0.0348} & {0.0106} \\
        (e) & \checkmark & \checkmark & \checkmark & \checkmark & 1.00 & {31.49} & {33.46} & {0.0375} & {0.0120} \\
        (f) & \checkmark & \checkmark & \checkmark & \checkmark & SVDS & \textbf{33.04} & \textbf{35.62} & \textbf{0.0342} & \textbf{0.0103} \\
        \bottomrule
    \end{tabular}
    }
    \vspace{-0.7mm}
\end{table*}

\begin{figure*}[t]
    \centering
    \includegraphics[width=\linewidth]{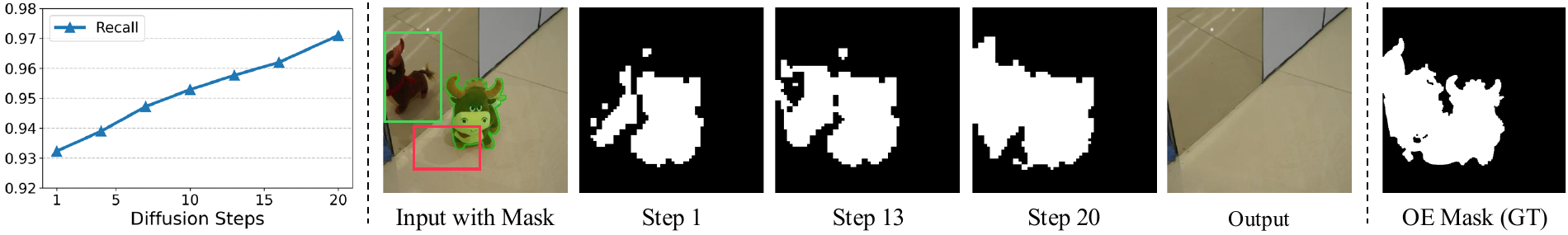}
    \vspace{-5mm}
    \caption{\textbf{Object-Effect Map from Adaptive Target-Aware Attention.} ObjectClear achieves a relatively high recall in the object-effect map, with recall values increasing as the denoising step progresses. The attention map obtained in the final step effectively covers both the target object and its associated effects, including the object’s shadow (\textcolor{red}{red} box) and its reflection in the mirror (\textcolor{green}{green} box).}
    \label{fig:attn_map}
    \vspace{-0.7mm}
\end{figure*}

\begin{figure*}[!t]
    \centering
    \includegraphics[width=\linewidth]{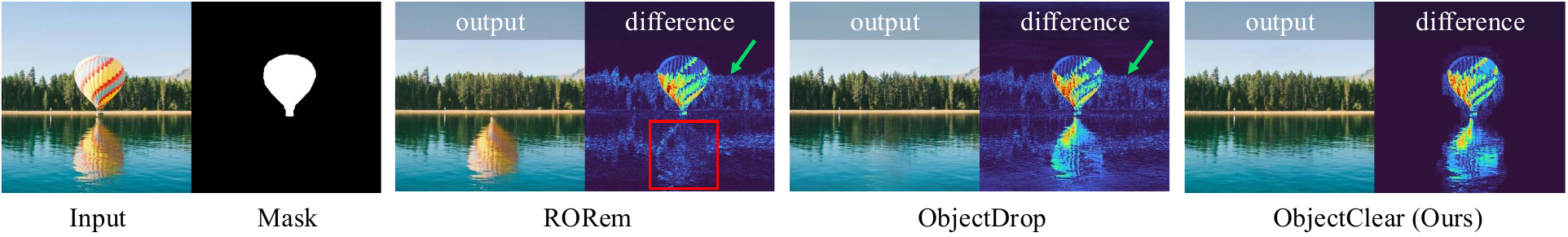}
    \vspace{-6mm}
    \caption{\textbf{Effectiveness of Attention-Guided Fusion.} The difference maps visualize the pixel-wise differences between input and output. RORem~\citep{li2025rorem} fails to remove the reflection and alters the background, yielding low differences in the reflection area and high values in the background.
    In contrast, both ObjectDrop~\citep{winter2024objectdrop} and our ObjectClear successfully remove the reflection, with ObjectClear better preserving the background as indicated by lower differences.
    }
    \vspace{-5mm}
    \label{fig:bg_diff}
\end{figure*}

\noindent \textbf{Qualitative Evaluation.} Qualitative results are shown in Fig.~\ref{fig:sota_gt} and Fig.~\ref{fig:sota_wild}. Generation-based inpainting approaches, such as SDXL-INP~\citep{podell2023sdxl}, and PowerPaint~\citep{zhuang2024powerpaint}, often generate new objects within the masked regions while fail to remove the effects of the removed objects. In contrast, previous object removal methods, such as Attentive Eraser~\citep{sun2025attentive}, DesignEdit~\citep{jia2024designedit}, and RORem~\citep{li2025rorem} demonstrate strong performance in removing the target object itself, but still fail to eliminate the associated effects. 
OmniEraser~\citep{wei2025omnieraser} demonstrates the ability to remove both objects and their effects, though it may still leave residual objects or effects, occasionally erase non-target regions, or introduce background changes.
Similarly, OmniPaint~\citep{yu2025omnipaint} maintains good background fidelity but may occasionally fail to fully remove the target object and its associated effects.
Benefiting from our training data and network designs, ObjectClear excels at accurately removing both the target object and its effects. Moreover, it robustly handles complex scenarios involving multiple occluding objects (see the last row of Fig.~\ref{fig:sota_wild}).

\vspace{-2mm}
\subsection{Ablation Studies}
\vspace{-1mm}
\noindent \textbf{The Effectiveness of Adaptive Target-Aware Attention (ATA).}
The ATA is designed to explicitly guide ObjectClear to attend to both the target object and its effects, leading to more complete and precise removal. From Table~\ref{tab:ablation} (a) to (b), improvements can be seen across all metrics with ATA introduced,
indicating its effectiveness in improving object removal accuracy (see visual comparison in supp).
Besides, as shown in the plot of Fig.~\ref{fig:attn_map} (left), the object-effect map from ATA module achieves a high recall of 0.97. In Fig.~\ref{fig:attn_map} (right), the attention map from the final denoising step (Step 20) effectively reveals object's shadow (\textcolor{red}{red} box) and mirror reflection (\textcolor{green}{green} box), altogether leading to complete and visually coherent object removal results.

\noindent \textbf{The Effectiveness of Simulated Data.}
The simulated data effectively enhances the scale and diversity of the training data, leading to notable improvements in object removal performance and better background preserving, as shown in Table~\ref{tab:ablation}(c). 
Our carefully designed simulation pipeline, especially the multi-object data synthesis, further enables ObjectClear to tackle challenging scenarios involving multiple objects. We provide more visual results in supplementary.

\noindent \textbf{Effectiveness of the Attention-Guided Fusion (AGF).}
The AGF is designed to preserve background consistency after object removal, a common challenge in generation-based methods like ObjectDrop~\citep{winter2024objectdrop} and RORem~\citep{li2025rorem}, as illustrated in Fig.~\ref{fig:bg_diff}. As the denoising process, our Adaptive Target-Aware Attention yields increasingly accurate attention maps (Fig.\ref{fig:attn_map}), which we leverage to guide the fusion of input background. This strategy significantly improves background preservation, as evidenced by large gains in PSNR-BG. Consequently, we observe a marked boost in overall performance, with PSNR increasing from $28.04$ to $32.77$, as shown in Table~\ref{tab:ablation}(c–d).

\noindent \textbf{Effectiveness of the Spatial-Varying Denoising Strength.}
The proposed Spatially-Varying Denoising Strength (SVDS) addresses two major issues: incomplete object removal or hallucinated objects when using $\mathrm{DS}=0.99$, and global color shifts when using $\mathrm{DS}=1.0$ (see visual ablation in supp.).
As shown in Table~\ref{tab:ablation}(d–f), SVDS (f) outperforms both $\mathrm{DS}=0.99$ (d) and $\mathrm{DS}=1.0$ (e). It effectively facilitates complete object removal while maintaining consistent background colors and details.

\section{Conclusion}
\label{sec:conclusion}
We introduce \textit{ObjectClear}, a practical framework for object removal that achieves high-quality object-effect removal while maintaining background consistency across diverse real-world scenarios. The framework employs an Adaptive Target-Aware Attention to adaptively focus on removal regions, along with Attention-Guided Fusion and Spatial-Varying Denoising Strength strategy to preserve background details. In addition, we present \textit{OBER}, a large-scale and diverse hybrid dataset that integrates camera-captured and simulated data. Thanks to our dataset and network designs, ObjectClear shows robust and superior performance, effectively overcoming key challenges in object removal and laying a solid foundation for future research in this field.

\clearpage

\vspace{2mm}
\noindent{\bf Acknowledgement.} This research is supported by cash and in-kind funding from NTU S-Lab and industry partner(s).

{
    \small
    \bibliographystyle{ieeenat_fullname}
    \bibliography{main}
}

\clearpage
\onecolumn

\begin{center}
	\Large\textbf{{Appendix}}\\
	\vspace{7mm}
\end{center}

In this appendix, we provide additional discussions and results to complete the main paper.
In Sec.~\ref{sec:additional_method_details}, we present further implementation details on data augmentation during training and other extended applications. 
In Sec.~\ref{sec:OBER_dataset_details}, we provide a more detailed introduction to the proposed OBER dataset, including statistics and some examples for demonstration.
In Sec.~\ref{sec:more_results}, we report additional results, such as further ablations, a user study, experiments with user strokes, results for object insertion and movement, and additional comparisons on in-the-wild images. In addition, we provide an online
\href{https://huggingface.co/spaces/jixin0101/ObjectClear}{[interactive demo]}
on Hugging Face, enabling users to remove arbitrary objects through simple clicks.

\section{Implementation Details}
\label{sec:additional_method_details}

\subsection{Training Augmentations}
During training, we apply on-the-fly random cropping to enable the network to learn object removal across varying object sizes. We also apply color augmentation and random flipping to enhance model robustness.
To improve the model's robustness to the estimated or user-provided coarse mask, several previous methods~\citep{winter2024objectdrop, jiang2025smarteraser, wei2025omnieraser} have introduced mask augmentation techniques. In line with these approaches, we also apply dilation and erosion to the input mask during training. Different from previous practice, our method employs an \textit{object-aware} dilation and erosion strategy, where the dilation and erosion kernel size is adaptively determined based on the size of the object.
As demonstrated in the qualitative results in Sec.~\ref{sec:user_input}, our method effectively handles coarse user-drawn masks by implicitly completing and refining them, showcasing strong robustness to imprecise inputs.

\subsection{More Details for Extended Applications}
\label{sec:other_app_details}
ObjectClear can be flexibly extended to various applications.
Visual results of object insertion and movement are shown in Fig.~\ref{fig:insert_repos}, where our models generate realistic visual effects.
In this section, we provide the implementation details of object insertion and object movement. 

\noindent \textbf{Object Insertion.}
The insertion network also leverages the OBER dataset for training and adopts the same architecture as ObjectClear, as illustrated in Fig.3 of the main paper,
which receives the input tuple of $<z_{t}, I_{in}, M_{o}, c>$ and supervised by $I_{GT}$ (for output image) and $M_{fg}$ (for Adaptive Target-Aware Attention). 
However, $I_{in}$, $I_{GT}$, and $c$ are constructed in a \textit{reverse manner} compared with the removal network.
Specifically, the ground-truth image $I_{GT}$ corresponds to the original image containing the object along with its associated effects, while $I_{in}$ is obtained by simply copying and pasting the object onto the background without any effects.
To generate $I_{in}$, we first extract the object from $I_{GT}$ using the object mask $M_{o}$ only, and then paste it into the corresponding background image.
In addition to the image pair, the input text $c$ is also reversed to ``\textit{insert the instance of}''. Notably, the Adaptive Target-Aware Attention map is still supervised by $M_{fg}$. Here, $M_{fg}$ refers to the mask that covers both the object and its \textit{generated} effects (\eg, shadows or reflections).

During inference, we obtain the input image $I_{in}$ by pasting an object foreground (w/o effects) onto the background image, and we feed it together with its corresponding object mask $M_{o}$ as the input pair. 
The network then generates the output image where both the object and its generated effects are harmoniously inserted.
Since the insertion network also integrates the Adaptive Target-Aware Attention, we further apply the Attention-Guided Fusion to preserve background fidelity while synthesizing realistic object effects.

\noindent \textbf{Object Movement.}
To enable object movement, we combine our object removal (ObjectClear) and insertion models into a two-stage framework. 
Specifically, we first apply ObjectClear to remove the target object and its associated effects, producing a clean, object-free background. 
The object is then extracted using its provided object mask, and users can specify a new location and adjust its scale before reinsertion. 
With our insertion network, the object is harmonized with the new context by generating realistic effects. 
Without retraining any additional networks, this two-stage pipeline supports controllable object movement while maintaining visual realism and consistency.

\section{More Details of OBER Dataset}
\label{sec:OBER_dataset_details}

\begin{figure*}[t]
	\centering
	\includegraphics[width=0.92\linewidth]{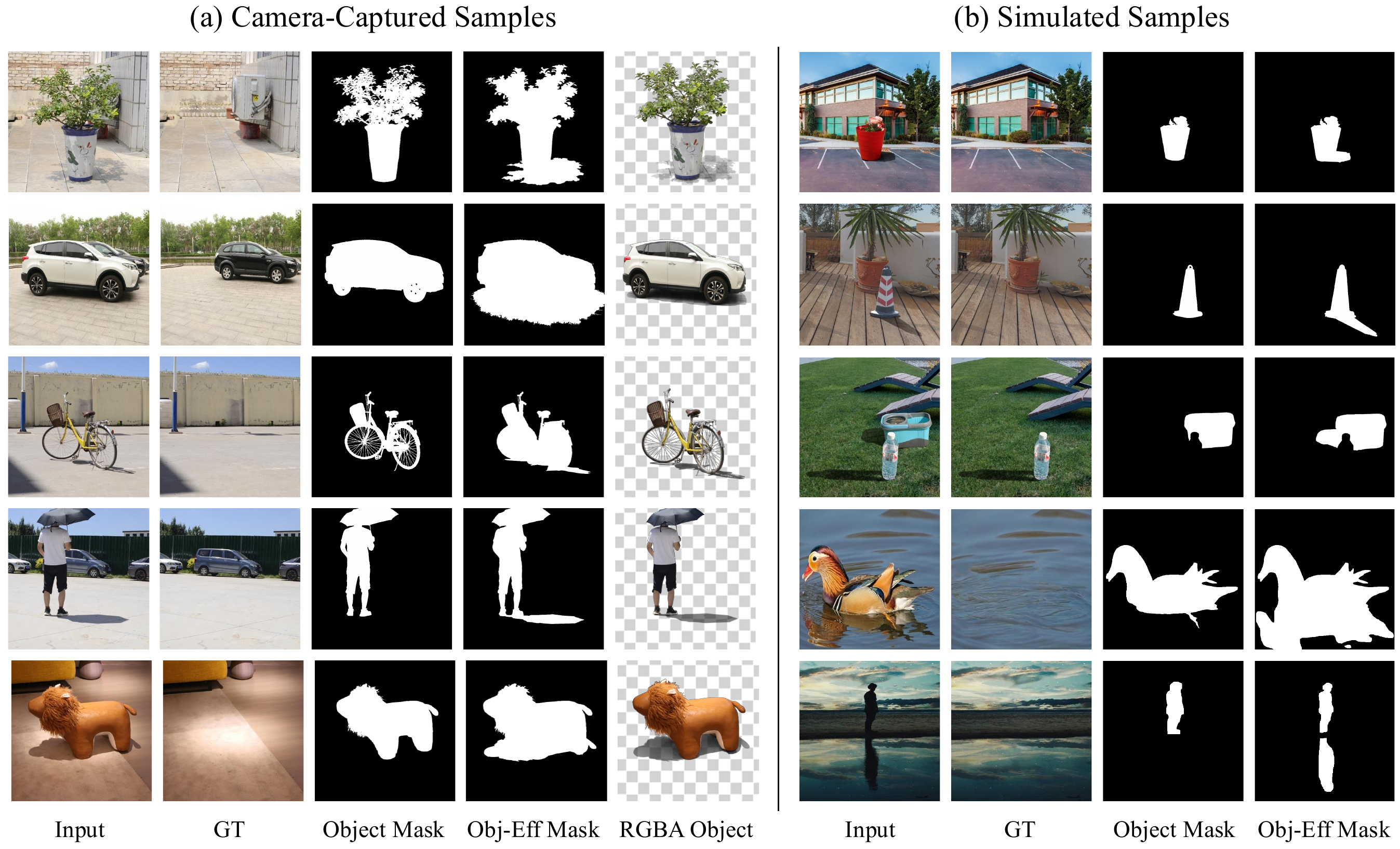}
	\vspace{-2mm}
	\caption{
		\textbf{Samples from our OBER Dataset.}
		We provide some samples from our OBER dataset including both camera-captured data and simulated data. While all data from both categories are annotated with fine-grained object masks and object-effect masks, we also extract RGBA foreground objects with associated effects from camera-captured data (as discussed in Sec.~3.1 in the main paper), which could be used to construct realistic simulated samples. In the simulated samples, we not only include the shadow effects but also the reflection effects, which are rarely included in other datasets, thanks to our reflection pair simulation strategy detailed in Sec.~\ref{sec:OBER_dataset_details}.
	}
	\label{fig:samples}
\end{figure*}

\begin{table}[t]
	\centering
	\caption{\textbf{Dataset Overview.} The OBER dataset consists of camera-captured and simulated training data, as well as two testing sets: OBER-Test (w/ ground truth) and OBER-Wild (w/o ground truth). All data subsets are annotated with object masks and object-effect masks.}
	\renewcommand{\arraystretch}{1.15}
	\renewcommand{\tabcolsep}{6.8mm}
	\resizebox{0.9\linewidth}{!}{
		\begin{tabular}{l|cc|cc}
			\toprule
			\multicolumn{1}{l|}{\multirow{2}{*}{\textbf{Properties}}} & \multicolumn{2}{c|}{\textbf{Training Set}} & \multicolumn{2}{c}{\textbf{Testing Set}} \\ 
			\cline{2-5}  & Camera-Captured & Simulated & OBER-Test & OBER-Wild \\
			\midrule \midrule
			{\#Image Pairs} & 2,715 & 10,000 & 163 & 302 \\
			{w/ Ground Truth} & \checkmark & \checkmark & \checkmark & × \\
			{w/ Object Mask} & \checkmark & \checkmark & \checkmark & \checkmark \\
			{w/ Object-Effect Mask} & \checkmark & \checkmark & \checkmark & \checkmark \\
			\bottomrule
		\end{tabular}
	}
	\label{tab:dataset_stats}
\end{table}

\noindent \textbf{Overview.}
Table~\ref{tab:dataset_stats} provides an overview of our OBER dataset \href{https://huggingface.co/datasets/sczhou/OBERDataset_ObjectClear}{[link]}. The training set consists of 2,715 camera-captured image pairs and 10,000 simulated pairs, all annotated with object masks and object-effect masks. For evaluation, we provide two test subsets: OBER-Test (includes 163 pairs with ground truth) and OBER-Wild (includes 302 pairs without ground truth), where object masks and object-effect masks are also available. 
Figure~\ref{fig:samples} presents samples from our OBER dataset, including camera-captured and simulated data with annotations such as object masks, object-effect masks, and RGBA foregrounds.

\noindent \textbf{Reflection Pair Simulation.}
Thanks to the strong priors of generative models, we observed that a model trained with only limited indoor mirror-reflection data can still generalize to removing simple outdoor reflection effects (\eg, water reflections), which is consistent with the findings in ObjectDrop~\citep{winter2024objectdrop}. 
However, the model often fails in more challenging cases, particularly when the object and its reflection are spatially separated or when the reflection is heavily distorted by water ripples. 
To address this limitation, we adopt a human-in-the-loop strategy to collect paired reflection data. 
Specifically, we first use our trained model to perform inference on 200 real-world reflection images, and then manually select 50 high-quality results (see examples in Fig.~\ref{fig:samples}(b)), which are then added as an important supplement to the training data. 
We found that even a small amount of high-quality reflection pairs can significantly improve the model's generalization ability across diverse reflection scenarios.

\noindent \textbf{Comparison with Existing Datasets.}
We compare our OBER dataset with existing datasets, including those focused on shadow removal (DESOBA-v2~\citep{liu2024shadow}) and object removal (RORD~\citep{Sagong2022rord}, MULAN~\citep{tudosiu2024mulan}, Counterfactual Dataset~\citep{winter2024objectdrop}, Video4Removal~\citep{wei2025omnieraser}). 
As summarized in Table~\ref{tab:dataset_comparison}, unlike prior datasets, OBER provides all three types of mask annotations as well as RGBA object foregrounds, enabling a wide range of tasks, including object removal, effect removal, and joint object–effect removal.
DESOBA-v2~\citep{liu2024shadow} targets only effect removal and thus does not handle the objects. 
In contrast, MULAN~\citep{tudosiu2024mulan} focuses solely on object removal without associated effects such as shadows or reflections. 
Moreover, the ground-truth images in both DESOBA-v2 and MULAN are produced by existing inpainting models, which may limit their visual realism.
RORD~\citep{Sagong2022rord} is among the first to provide object–effect masks, however, these masks are \textit{very coarse} and can not separately annotate objects and their effects. 
Consequently, it cannot support independent removal of either the object or the effect. 
The Counterfactual Dataset~\citep{winter2024objectdrop} and Video4Removal Dataset~\citep{wei2025omnieraser} are designed for object–effect removal tasks, the same with ours, but they are not publicly accessible. 
Furthermore, these datasets provide only object masks, lacking effect masks and RGBA foregrounds, which limits their scalability and diverse usage.
In contrast, our OBER dataset provides richer annotations, including precise and separate masks for objects and their effects, together with RGBA foregrounds, enabling more flexible and fine-grained removal tasks. 
Notably, the effect masks serve as crucial supervision for learning accurate object–effect removal while preserving background fidelity. 
OBER is a hybrid dataset comprising high-quality captured and realistic synthetic data, covering diverse and complex scenarios such as multi-object occlusions and indoor/outdoor reflections.
\textbf{We will release our OBER dataset publicly}, which we believe will significantly benefit future research in this field.

\begin{table*}[t]
	\centering
	\caption{\textbf{Comparison of OBER Dataset with Existing Datasets.} $*$ indicates that the dataset is not publicly available.}
	\renewcommand{\arraystretch}{1.25}
	\renewcommand{\tabcolsep}{1.5mm}
	\resizebox{\linewidth}{!}{
		
		\begin{tabular}{@{}c|l|ccccccc}
			\toprule
			\makecell{} & \makecell{\textbf{Description}} & \makecell{\textbf{RORD}~\citep{Sagong2022rord}} & \makecell{\textbf{MULAN}~\citep{tudosiu2024mulan}} & \makecell{\textbf{DESOBAv2}~\citep{liu2024shadow}} & \makecell{\textbf{Counterfactual$^{*}$}~\citep{winter2024objectdrop}} & \makecell{\textbf{Video4Removal$^{*}$}~\citep{wei2025omnieraser}} & \makecell{\textbf{OBER (ours)}} \\
			\midrule
			\multirow{3}{*}{\rotatebox{90}{Tasks}} 
			& Object Removal        & ×          & \checkmark & ×          & \checkmark & \checkmark & \checkmark \\
			& Effect Removal        & ×          & ×          & \checkmark & \checkmark & \checkmark & \checkmark \\
			& Object-Effect Removal & \checkmark & ×          & ×          & \checkmark & \checkmark & \checkmark \\
			\midrule
			\multirow{6}{*}{\rotatebox{90}{Annotations}} 
			& Object Mask           & ×          & \checkmark & \checkmark & \checkmark & \checkmark & \checkmark \\
			& Effect Mask           & ×          & ×          & \checkmark & ×          & ×          & \checkmark \\
			& Object-Effect Mask    & \checkmark (Coarse) & × & \checkmark & ×          & ×          & \checkmark \\
			\cline{2-8}
			& RGBA Objects          & ×          & ×          & ×          & ×          & ×          & \checkmark \\
			& Multi Objects         & \checkmark & \checkmark & ×          & \checkmark & \checkmark & \checkmark \\
			\cline{2-8}
			& Camera-Captured GT    & \checkmark & ×          & ×          & \checkmark & \checkmark & \checkmark \\
			\bottomrule
		\end{tabular}
	}
	\vspace{-2mm}
	\label{tab:dataset_comparison}
\end{table*}

\section{More Results}
\label{sec:more_results}

\subsection{Results for Ablation Study}

\begin{figure*}[h]
	\centering
	\includegraphics[width=0.92\linewidth]{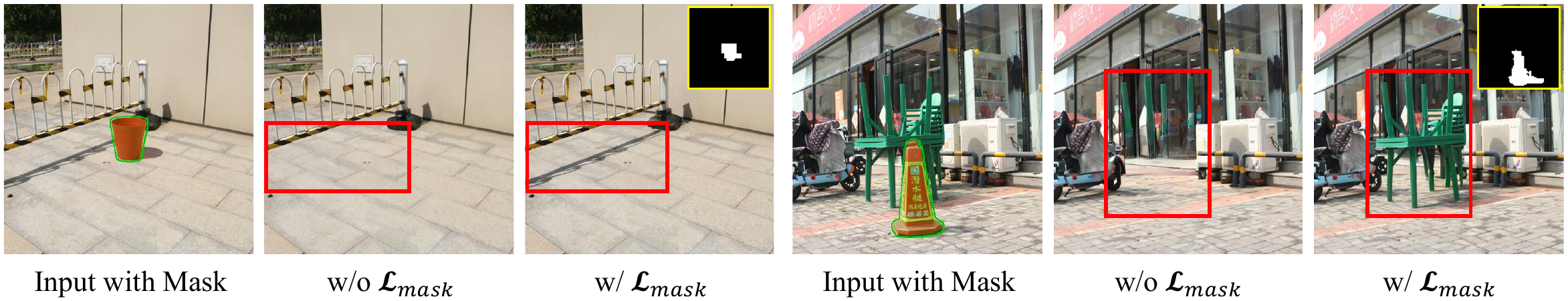}
	\vspace{-1mm}
	\caption{
		\textbf{Effectiveness of Adaptive Target-Aware Attention (ATA).}
		It could be observed that when training without ATA (w/o $\mathcal{L}_{mask}$), the model struggles to accurately remove the target object and its effects, leading to mistakenly erasing unrelated background object (\textit{right}) or effects (\textit{left}). In contrast, when introducing TAT trained with the mask supervision $\mathcal{L}_{mask}$, the attention maps (shown in yellow boxes) can accurately localize the removal regions, leading to more precise and complete removal results.
	}
	\label{fig:ablation_mask_loss}
	\vspace{-1mm}
\end{figure*}

\textbf{Effectiveness of Adaptive Target-Aware Attention.}
Thanks to the OBER dataset’s rich annotations, Adaptive Target-Aware Attention (ATA) leverages the object-effect mask and its supervision loss  $\mathcal{L}_{mask}$. It guides cross-attention layers to focus on the object and its associated effects while preserving background textures, which enables decoupled optimization of object removal and background reconstruction. As shown in Fig.~\ref{fig:ablation_mask_loss}, ATA adaptively identifies object-effect regions to be removed, as reflected in the attention mask (shown in yellow boxes). This leads to more accurate and complete removal of objects and their shadow effects, without mistakenly erasing unrelated background content.

\begin{figure*}[t]
	\centering
	\includegraphics[width=0.92\linewidth]{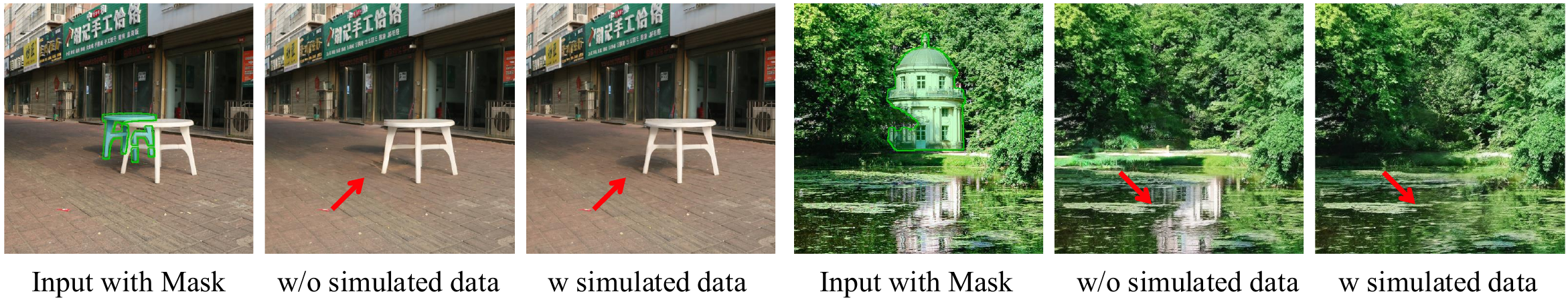}
	\vspace{-2mm}
	\caption{
		\textbf{Effectiveness of Simulated Data.}
		Since our simulation data includes multi-object compositions, training with such data enables the model to accurately remove the target object and its associated effects while preserving unrelated object effects (\textit{left}). In addition, adding the reflection data pairs during training greatly enhances the model capability of removing reflections, even in challenging cases (\textit{right}).
	}
	\label{fig:ablation_simulated_data}
	\vspace{-1mm}
\end{figure*}

\begin{figure*}[!t]
	\centering
	\includegraphics[width=0.92\linewidth]{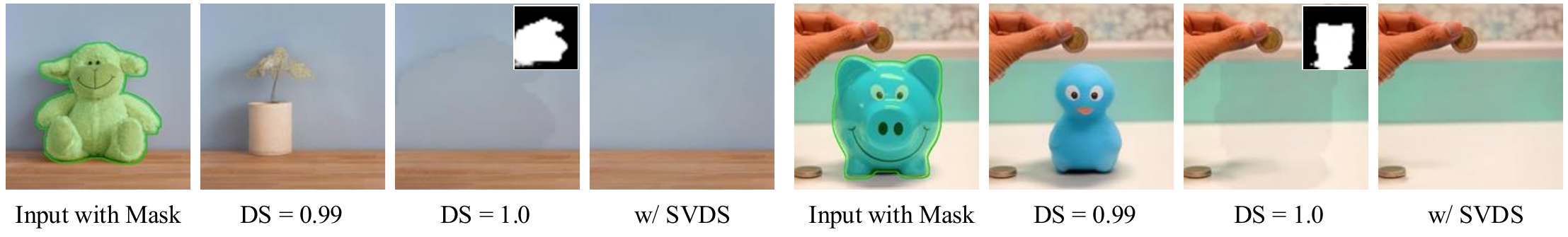}
	\vspace{-2mm}
	\caption{
		\textbf{Effectiveness of Spatially-Varying Denoising Strength (SVDS).} $\mathrm{DS}=0.99$ often leads to incomplete removal or hallucinated objects, while $\mathrm{DS}=1.0$ causes noticeable color inconsistency (shown after AGF, where the background is from the input image and the object/affected areas are from the removal results). In contrast, SVDS achieves complete object removal with consistent background colors.
	}
	\label{fig:ablation_svds}
	\vspace{-1mm}
\end{figure*}

\begin{figure*}[t]
	\centering
	\includegraphics[width=0.92\linewidth]{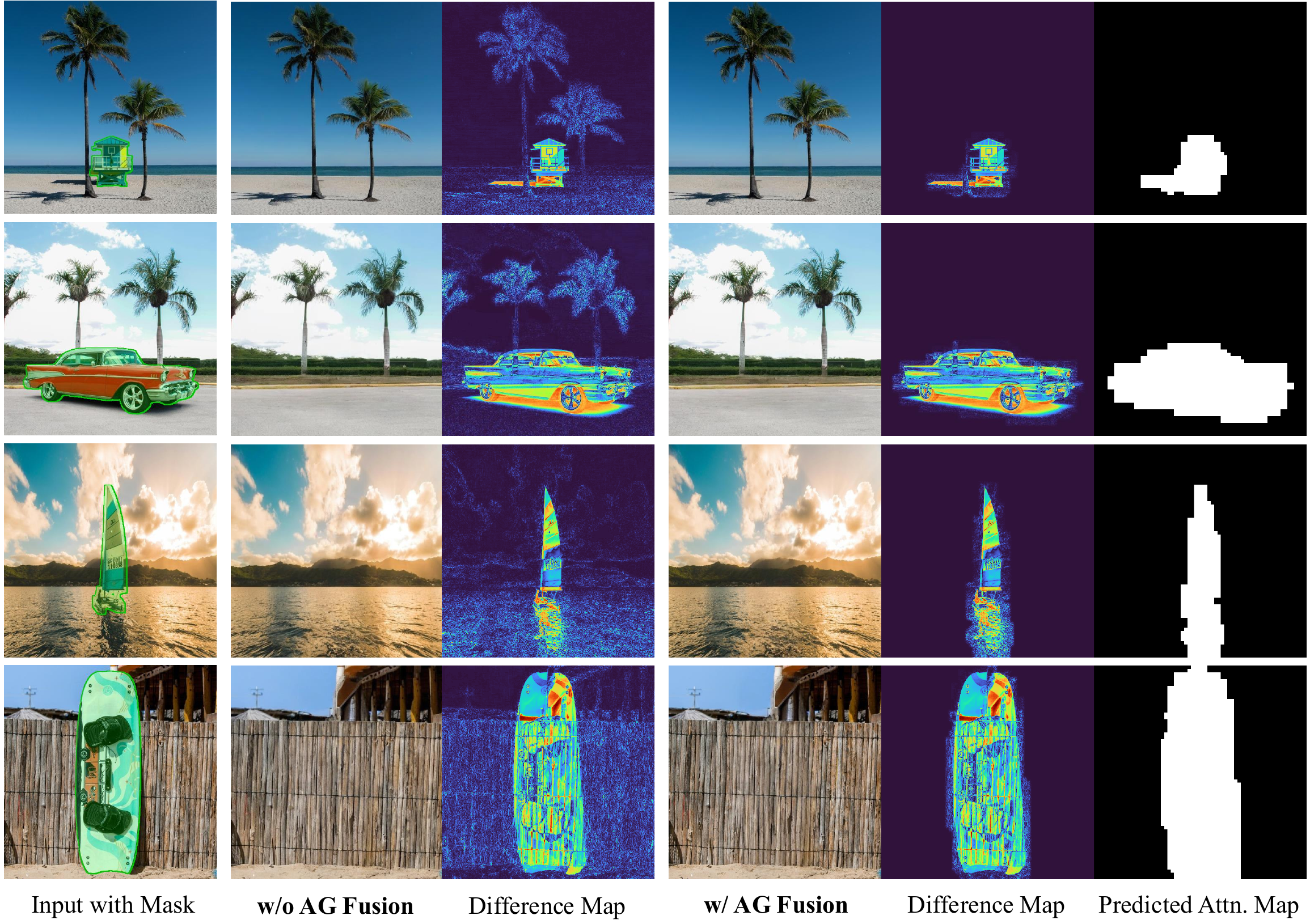}
	\vspace{-2mm}
	\caption{
		\textbf{Effectiveness of Attention-Guided Fusion (AG Fusion).}    
		We visualize background detail changes by computing difference maps between the generated image and input. Ideally, these maps should exhibit large values only within the object removal regions, covering the object and its effects, while remaining minimal in the unrelated background areas. As observed, without AG Fusion, noticeable differences appear in the background. In contrast, with the AG Fusion strategy, these undesired background discrepancies are largely eliminated.
	}
	\label{fig:ablation_fusion}
	\vspace{-5mm}
\end{figure*}

\noindent \textbf{Effectiveness of Simulated Data.}
To balance realism and scalability, in addition to camera-captured data, we augment OBER with a simulation pipeline. Our simulation data is generated by compositing the RGBA foreground objects (extracted from camera-captured data) onto diverse backgrounds. In particular, the simulation of multi-object compositions leads to notable improvements in object removal robustness under mutual occlusions, as shown in Fig.~\ref{fig:ablation_simulated_data} (\textit{left}). Furthermore, our simulated reflection pairs greatly improve the model capability of removing reflections, even in challenging cases (Fig.~\ref{fig:ablation_simulated_data}, \textit{right}).

\noindent \textbf{Effectiveness of Attention-Guided Fusion.}
The attention map supervised by $\mathcal{L}_{mask}$ supports the Attention-Guided Fusion (AG Fusion) strategy during inference. It helps to blend the generated image with the original input via a copy-and-paste operation, where pixels within the object-effect region are taken from the generated result, and the rest are preserved from the original image. Such practice effectively reduces undesired background detail changes caused by VAE reconstruction errors and the diffusion process, thereby greatly preserving the background fidelity. In Fig.~\ref{fig:ablation_fusion}, we visualize the background detail changes by showing the difference maps between the generated image and the corresponding input, where a clear improvement on background preservation could be observed.

\noindent \textbf{Effectiveness of Spatially-Varying Denoising Strength.}
In diffusion-based image editing, the initial latent for denoising is typically obtained by adding noise to the input image latent. The denoising strength (DS), $\mathrm{DS} \in [0,1]$, controls the noise level: a larger value injects more noise, thereby pushing the initial noisy latent closer to the pure-noise prior. When $\mathrm{DS}=1.0$, the diffusion process starts entirely from noise, discarding information from the input. In this paper, we propose Spatial-Varying Denoising Strength (SVDS), which applies $\mathrm{DS}=1.0$ within the masked object region and $\mathrm{DS}=0.99$ (an empirical setting commonly adopted by previous methods~\citep{podell2023sdxl}) outside in the unmasked background, ensuring complete object removal while maintaining color consistency. As shown in Fig.~\ref{fig:ablation_svds}, setting $\mathrm{DS}=0.99$ often leads to incomplete removal or hallucinated objects, whereas $\mathrm{DS}=1.0$ results in noticeable color inconsistency. To highlight this inconsistency, the results of $\mathrm{DS}=1.0$ in Fig.~\ref{fig:ablation_svds} are obtained after the Attention-Guided Fusion (AGF) operation, where the background is taken from the original input image, while the object and affected areas are taken from the removal results. In contrast, our method with SVDS achieves superior performance in both object removal and preservation of background color consistency.

\begin{table*}[t]
	\centering
	\vspace{-1mm}
	\caption{\textbf{Quantitative comparison on OmniPaint-Bench and OmniEraser-Bench.} The best and second performances are marked in \colorbox{rred}{\underline{red}} and \colorbox{oorange}{orange}, respectively.}
	\vspace{-1mm}
	\renewcommand{\arraystretch}{1.15}
	\renewcommand{\tabcolsep}{4.0mm}
	\resizebox{\textwidth}{!}{
		\begin{tabular}{lcccccccc}
			\toprule
			\multirow{2}{*}{Methods} & \multicolumn{4}{c}{OmniPaint-Bench} & \multicolumn{4}{c}{OmniEraser-Bench} \\
			\cmidrule(lr){2-5} \cmidrule(lr){6-9}
			& PSNR$\uparrow$ & LPIPS$\downarrow$ & DINO$\downarrow$ & CLIP$\downarrow$
			& PSNR$\uparrow$ & LPIPS$\downarrow$ & DINO$\downarrow$ & CLIP$\downarrow$ \\
			\midrule
			SDXL-INP~\cite{podell2023sdxl}             & 20.51 & 0.1924 & 0.0929 & 0.1100 & 21.26 & 0.1968 & 0.0938 & 0.0926 \\
			PowerPaint~\cite{zhuang2024powerpaint}     & 21.55 & 0.1882 & 0.0874 & 0.0990 & 22.21 & 0.2001 & 0.0911 & 0.0907 \\
			BrushNet~\cite{ju2024brushnet}             & 19.00 & 0.2733 & 0.1909 & 0.1446 & 20.58 & 0.2098 & 0.1182 & 0.1098 \\
			DesignEdit~\cite{jia2024designedit}        & 24.32 & 0.1763 & 0.0769 & 0.0717 & 23.87 & 0.2168 & 0.1020 & 0.0755 \\
			CLIPAway~\cite{ekin2024clipaway}           & 20.07 & 0.1985 & 0.0929 & 0.1129 & 20.78 & 0.2035 & 0.0934 & 0.0956 \\
			FreeCompose~\cite{chen2024freecompose}     & 22.25 & 0.1631 & 0.0723 & 0.0770 & 22.60 & 0.1782 & 0.0733 & 0.0688 \\
			Attentive Eraser~\cite{sun2025attentive}   & 24.17 & 0.1523 & 0.0501 & 0.0590 & 24.77 & 0.1538 & 0.0463 & 0.0388 \\
			RORem~\cite{li2025rorem}                   & 23.92 & 0.1462 & 0.0478 & 0.0576 & 23.70 & 0.1746 & 0.0532 & 0.0416 \\
			OmniEraser~\citep{wei2025omnieraser}       & 23.02 & 0.2034 & 0.0564 & 0.0640 & 23.83 & 0.1766 & 0.0460 & 0.0481 \\
			OmniPaint~\citep{yu2025omnipaint}          & \colorbox{oorange}{25.56} & \colorbox{oorange}{0.1001} & \colorbox{oorange}{0.0249} & \colorbox{oorange}{0.0367} & \colorbox{oorange}{25.67} & \colorbox{oorange}{0.1069} & \colorbox{oorange}{0.0213} & \colorbox{oorange}{0.0182} \\
			\textbf{ObjectClear (Ours)}                & \colorbox{rred}{\underline{26.35}} & \colorbox{rred}{\underline{0.0950}} & \colorbox{rred}{\underline{0.0244}} & \colorbox{rred}{\underline{0.0334}} 
			& \colorbox{rred}{\underline{27.90}} & \colorbox{rred}{\underline{0.0942}} & \colorbox{rred}{\underline{0.0230}} & \colorbox{rred}{\underline{0.0142}} \\
			\bottomrule
	\end{tabular}}
	\vspace{-2mm}
	\label{tab:object_effect_remove_benchmark}
\end{table*}

\begin{table*}[t]
	\centering
	\caption{\textbf{Quantitative comparison on MULAN Dataset.} The best and second performances are marked in \colorbox{rred}{\underline{red}} and \colorbox{oorange}{orange}, respectively.}
	\renewcommand{\arraystretch}{1.2}
	\renewcommand{\tabcolsep}{4.0mm}
	\resizebox{0.6\textwidth}{!}{
		\begin{tabular}{lcccc}
			\toprule
			Methods & PSNR$\uparrow$ & LPIPS$\downarrow$ & DINO$\downarrow$ & CLIP$\downarrow$ \\
			\midrule
			SDXL-INP~\cite{podell2023sdxl}           & 19.91 & 0.2494 & 0.1324 & 0.1312 \\
			PowerPaint~\cite{zhuang2024powerpaint}   & 21.18 & 0.2449 & 0.1087 & 0.0962 \\
			BrushNet~\cite{ju2024brushnet}           & 18.22 & 0.3181 & 0.2062 & 0.1893 \\
			DesignEdit~\cite{jia2024designedit}      & 23.26 & 0.2375 & 0.1114 & 0.0725 \\
			CLIPAway~\cite{ekin2024clipaway}         & 20.08 & 0.2666 & 0.1180 & 0.1152 \\
			FreeCompose~\cite{chen2024freecompose}   & 21.30 & 0.2337 & 0.0828 & 0.0703 \\
			Attentive Eraser~\cite{sun2025attentive} & \colorbox{oorange}{23.96} & 0.1960 & \colorbox{oorange}{0.0551} & \colorbox{oorange}{0.0397} \\
			RORem~\cite{li2025rorem}                 & 23.53 & 0.2369 & 0.0571 & 0.0438 \\
			OmniEraser~\citep{wei2025omnieraser}           & 21.56 & 0.2642 & 0.0728 & 0.0682 \\
			OmniPaint~\citep{yu2025omnipaint}        & 22.29 & \colorbox{oorange}{0.1915} & 0.0650 & 0.0550 \\
			\textbf{ObjectClear (Ours)}        & \colorbox{rred}{\underline{24.89}} & \colorbox{rred}{\underline{0.1586}} & \colorbox{rred}{\underline{0.0468}} & \colorbox{rred}{\underline{0.0373}} \\
			\bottomrule
	\end{tabular}}
	\label{tab:mulan_benchmark}
\end{table*}

\subsection{Comparisons on Additional Benchmarks}
To further evaluate the robustness and generalization ability of our method, we conduct additional evaluations on three benchmarks, \ie, OmniEraser-Bench~\citep{wei2025omnieraser}, OmniPaint-Bench~\citep{yu2025omnipaint}, MULAN~\citep{tudosiu2024mulan}.

\noindent\textbf{OmniEraser-Bench and OmniPaint-Bench Datasets.}
OmniEraser-Bench~\citep{wei2025omnieraser} and OmniPaint-Bench~\citep{yu2025omnipaint} are two recent benchmarks for object–effect removal, which align with our task setting. 
For a fair comparison, all methods use their default input sizes (OmniEraser~\citep{wei2025omnieraser} at 1024, all others at 512), then resize the outputs to the same size (short side 512) for evaluation. Table~\ref{tab:object_effect_remove_benchmark} shows our approach outperforms all baselines across metrics on both benchmarks.

\noindent\textbf{MULAN Dataset.}
Since the ground truth of some samples in MULAN~\citep{tudosiu2024mulan} retains shadows or reflections, it is not suitable for evaluating object–effect removal. 
Therefore, we randomly sample 500 ``effect-free'' image pairs from MULAN for our evaluation. 
As shown in Table~\ref{tab:mulan_benchmark}, our method achieves the best performance across all metrics, even surpassing RORem~\citep{li2025rorem}, which is trained on MULAN, demonstrating the strong object removal capability of our approach. 
This also confirm that our model performs robustly on ``effect-free'' object removal datasets, without introducing negative effects when the objects do not show any shadows or reflections.

\subsection{Fair Comparison with Object-Effect Mask}
Our Attention-Guided Fusion (AGF) module leverages object-effect masks predicted by the proposed Adaptive Target-Aware Attention to blend the original input background back into the generated result. Importantly, these masks are predicted by our model rather than taken from annotations, ensuring that we do not use any privileged information unavailable to other approaches. Furthermore, existing baseline methods do not have the capability to predict object–effect masks and therefore cannot perform background blending in the same way. This makes AGF an integral part of our model design rather than an external post-processing step, and the comparisons in the main paper are therefore fair.

To further demonstrate that the performance gain is not solely due to the background blending, we perform an additional experiment where all baseline methods are given the ground-truth object-effect mask for blending. As shown in Table~\ref{tab:blended_results}, our method continues to outperform all baselines under this setting, indicating that our superior results primarily come from more effective object–effect generation rather than the availability of blending masks.

\begin{table*}[h]
	\centering
	\caption{\textbf{Quantitative results on RORD-Val with object-effect masks for blending.} All baseline methods are equipped with background blending using ground-truth object-effect masks. The best and second performances are marked in \colorbox{rred}{\underline{red}} and \colorbox{oorange}{orange}, respectively. Our method using our predicted object-effect masks achieves the best performance across all metrics.}
	\renewcommand{\arraystretch}{1.2}
	\renewcommand{\tabcolsep}{4.0mm}
	\resizebox{0.65\textwidth}{!}{
		\begin{tabular}{l|cccc}
			\toprule
			Method & PSNR$\uparrow$ & LPIPS$\downarrow$ & DINO$\downarrow$ & CLIP$\downarrow$ \\
			\midrule
			SDXL-INP~\cite{podell2023sdxl} (w/b)           & 21.67 & 0.1592 & 0.0688 & 0.0932 \\
			PowerPaint~\cite{zhuang2024powerpaint} (w/b)   & 22.51 & 0.1472 & 0.0560 & 0.0606 \\
			BrushNet~\cite{ju2024brushnet} (w/b)           & 18.48 & 0.2421 & 0.1572 & 0.1465 \\
			DesignEdit~\cite{jia2024designedit}  (w/b)     & 23.73 & 0.1580 & 0.0721 & 0.0755 \\
			CLIPAway~\cite{ekin2024clipaway}  (w/b)        & 21.96 & 0.1735 & 0.0666 & 0.0734 \\
			FreeCompose~\cite{chen2024freecompose} (w/b)   & 23.08 & 0.1603 & 0.0829 & 0.0834 \\
			Attentive Eraser~\cite{sun2025attentive}  (w/b)& 22.90 & 0.1700 & 0.0880 & 0.0854 \\
			RORem~\cite{li2025rorem}  (w/b)                & \colorbox{oorange}{25.24} & \colorbox{oorange}{0.1398} & \colorbox{oorange}{0.0387} & \colorbox{oorange}{0.0498} \\
			\textbf{ObjectClear (Ours)}          &  \colorbox{rred}{\underline{26.24}} & \colorbox{rred}{\underline{0.1157}} & \colorbox{rred}{\underline{0.0191}} & \colorbox{rred}{\underline{0.0299}} \\
			\bottomrule
		\end{tabular}
	}
	\label{tab:blended_results}
\end{table*}

\subsection{Generalization to Multi-Object Removal}
\begin{figure*}[h]
	\centering
	\includegraphics[width=0.92\linewidth]{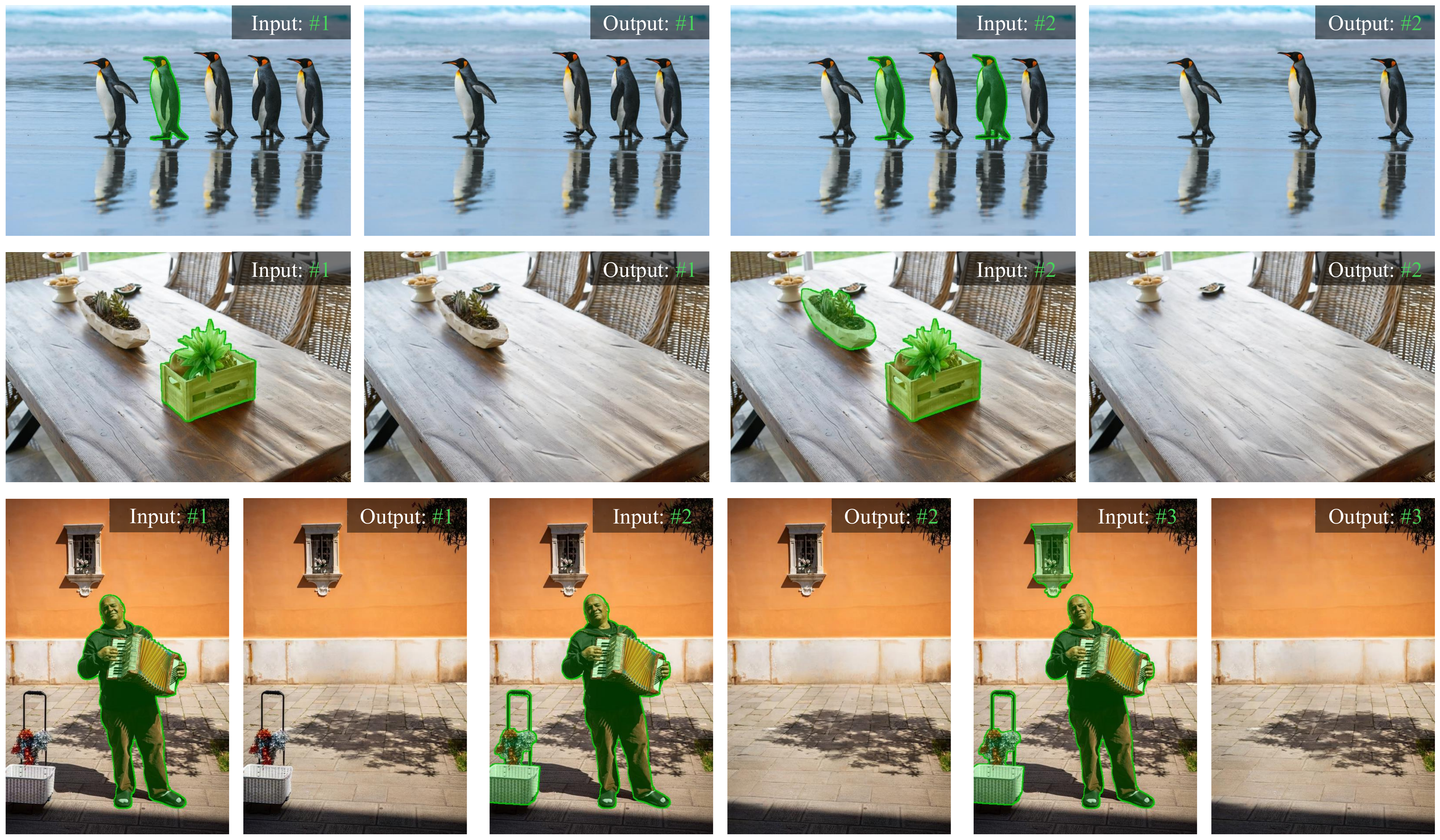}
	\caption{
		\textbf{Multi-object removal.} Our method removes one or multiple objects simultaneously using masks to indicate the targets.
	}
	\label{fig:multi_object_removal}
\end{figure*}

The OBER dataset also covers multi-object removal cases, and our model generalizes well to such scenarios (Fig.~\ref{fig:multi_object_removal}). 
This reflects the strong generalization capability of a network trained on our dataset.

\begin{figure}[!t]
	\centering
	\includegraphics[width=0.75\linewidth]{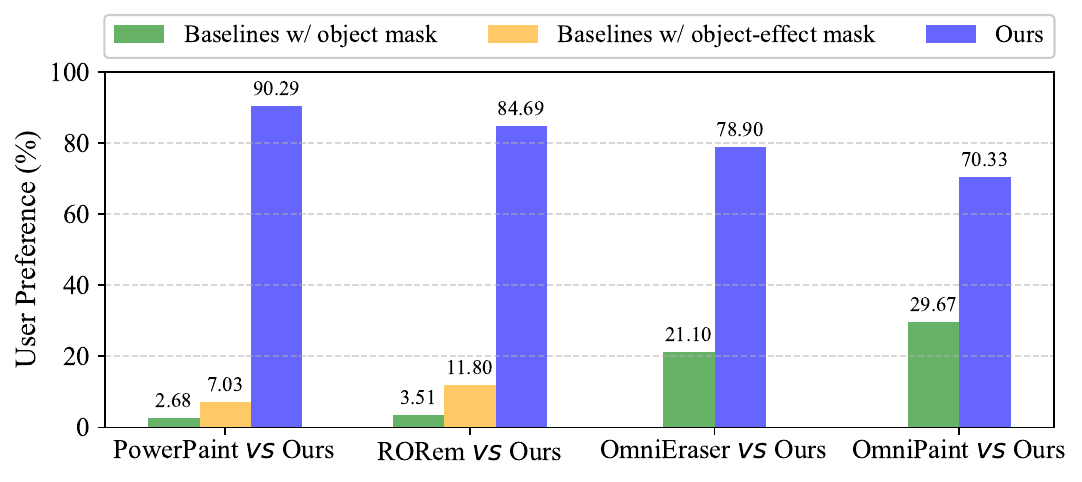}
	\vspace{-2mm}
	\caption{\textbf{User Study.} Our ObjectClear is preferred by human voters over four representative state-of-the-art methods (PowerPaint~\citep{zhuang2024powerpaint}, RORem~\citep{li2025rorem}, OmniEraser~\citep{wei2025omnieraser}, and OmniPaint~\citep{yu2025omnipaint}. 
	}
	\label{fig:user_study}
\end{figure}

\subsection{User Study}
To enable a more comprehensive evaluation, we conducted a user study on object removal results for in-the-wild images. We compared ObjectClear with four representative state-of-the-art methods: PowerPaint~\citep{zhuang2024powerpaint}, RORem~\citep{li2025rorem}, OmniEraser~\citep{wei2025omnieraser}, and OmniPaint~\citep{yu2025omnipaint}. 
For a fair comparison, we evaluated PowerPaint and RORem under two settings: (1) conditioned on the object mask and (2) conditioned on the object-effect mask, OmniEraser and OmniPaint are conditioned only on the object mask. All ObjectClear results were generated using only the object mask, and we compared our outputs against both settings of PowerPaint and RORem, as well as the object-mask setting of OmniEraser and OmniPaint.
We invite a total of 30 participants for this user study. Each volunteer was presented 80 randomly selected image quadruples, consisting of: \textit{an input image}, \textit{two results} from a baseline method under different mask conditions, and \textit{our result} (for OmniEraser and OmniPaint, only the object-mask result was provided, consistent with our setting, thus forming a triple set). Participants were asked to select the best removal result based on two criteria: the realism of the object region and the preservation of background details.
As summarized in Fig~\ref{fig:user_study}, ObjectClear outperforms the baselines under both mask settings. Notably, although some baseline methods benefited from access to object-effect masks, our ObjectClear won more user preference with the object mask only.

\subsection{Results with User Strokes}
\label{sec:user_input}

\begin{figure*}[h]
	\centering
	\includegraphics[width=0.92\linewidth]{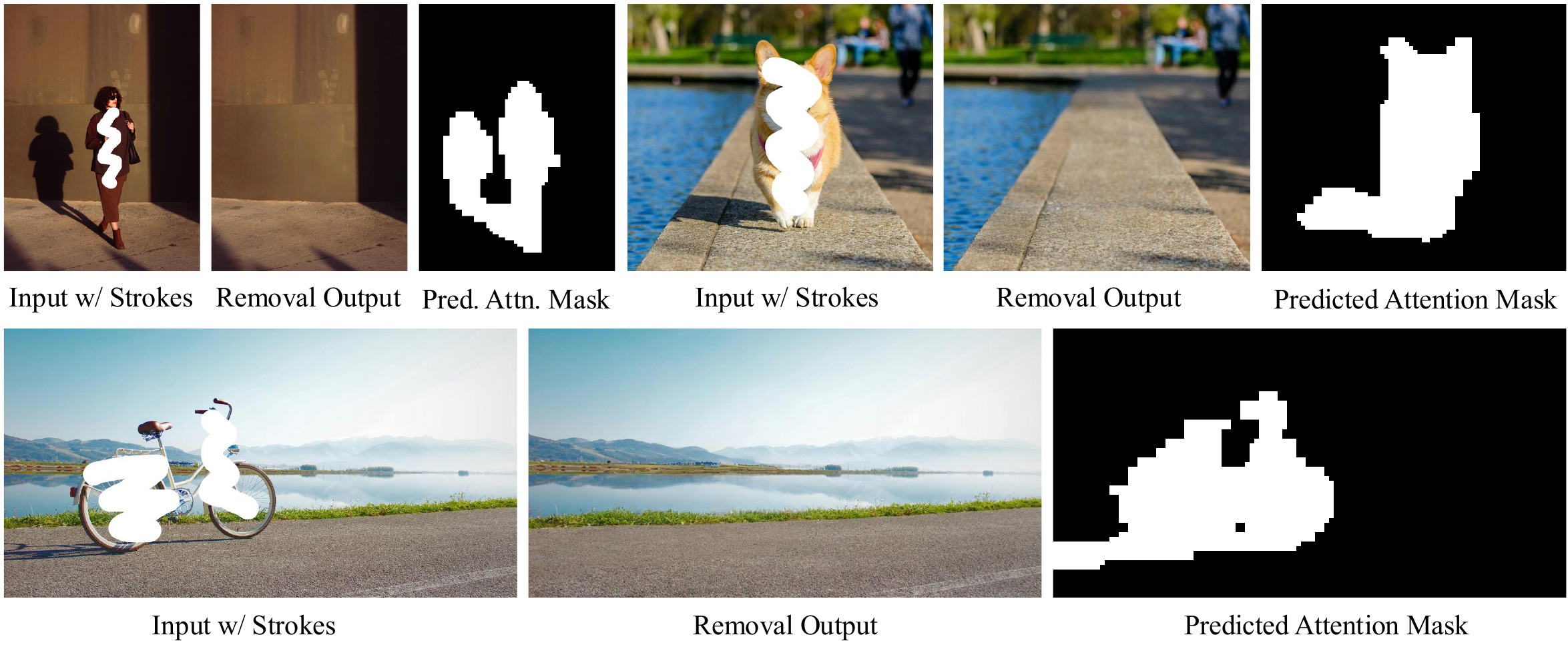}
	\vspace{-2mm}
	\caption{
		\textbf{Results with User Strokes.}
		We simulate user strokes and feed them into the network along with the images. The resulting outputs and attention maps show that our network can accurately identify and attend to the object and its effects, even with imprecise masks.
	}
	\label{fig:user_strokes}
\end{figure*}

In practical applications, users often interact with visual systems through imprecise or casually drawn inputs, such as rough scribbles or incomplete masks. These inputs may vary significantly in shape, location, and accuracy. Therefore, it is essential for a robust object removal network to effectively process such arbitrary mask inputs without relying on carefully crafted annotations. Benefiting from our mask augmentation strategy and Adaptive Target-Aware Attention mechanism, our network demonstrates strong robustness to diverse mask inputs. In this subsection, we simulate user strokes and feed them into the network along with the images. The resulting outputs and attention maps show that our network can accurately identify and attend to the object and its associated effects, even with imprecise masks, as illustrated in Fig.~\ref{fig:user_strokes}.

\subsection{Object Insertion and Movement.}
\begin{figure*}[th]
	\centering
	\includegraphics[width=\linewidth]{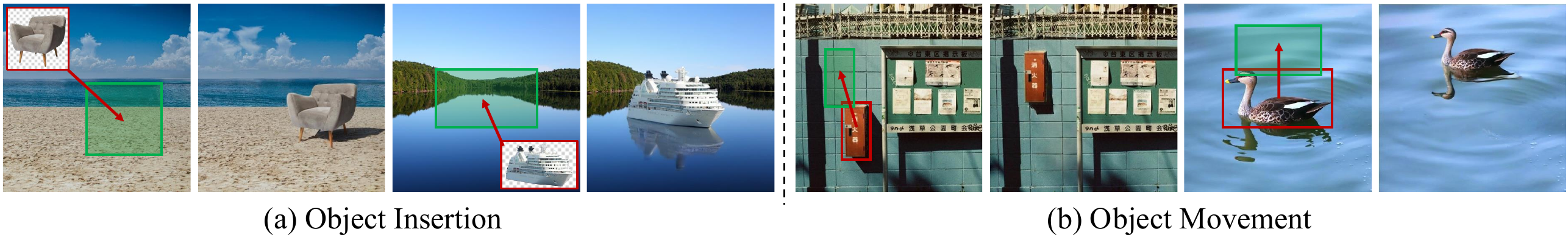}
	\vspace{-5mm}
	\caption{\textbf{Object Insertion and Movement.} In addition to accurately inserting or repositioning objects, our ObjectClear also generates plausible and natural shadows and reflections accordingly.}
	\label{fig:insert_repos}
\end{figure*}

As shown in Fig.~\ref{fig:insert_repos}, even when only the target objects are specified for insertion and movement, ObjectClear is capable of generating plausible and natural shadows and reflections accordingly.

\subsection{More Comparisons on In-the-wild Data}
\label{sec:in-the-wild_comparison}
In this subsection, we conduct comprehensive comparisons of ObjectClear with state-of-the-art methods across two categories: \textit{image inpainting methods} and \textit{object removal methods}, on the in-the-wild data.

\noindent \textbf{Comparison with Image Inpainting Methods.} We compare our method with state-of-the-art image inpainting methods, including PowerPaint~\citep{zhuang2024powerpaint}, Attentive Eraser~\citep{sun2025attentive}, DesignEdit~\citep{jia2024designedit}, and RORem~\citep{li2025rorem}. Since these inpainting methods often struggle to remove object-associated effects when only provided with an object mask, we supply them with our annotated Object-Effect Masks for fair comparison, while our ObjectClear uses only Object Masks as input. As shown in Fig.~\ref{fig:supp_object_removal_diff}, although these inpainting methods can remove shadows with the guidance of additional effect regions, they frequently introduce undesirable changes to the original background. In contrast, ObjectClear effectively eliminates the target object and its associated effects using only the object mask, while preserving background content with high fidelity.

\noindent \textbf{Comparison with Object Removal Methods.}
We further compare ObjectClear with object removal methods (OmniEraser~\citep{wei2025omnieraser} and OmniPaint~\citep{yu2025omnipaint}) under a unified input setting, all methods use Object Masks as input to ensure fairness. As illustrated in Fig.~\ref{fig:supp_object_removal}, our ObjectClear demonstrates superior stability compared to the two baselines: it not only successfully eliminates the target object and its associated effects but also avoids unintended changes to the original background, preserving background fidelity with high precision.
To further highlight the background preservation advantage, we visualize the pixel-wise difference maps of the compared methods in Fig.~\ref{fig:supp_object_removal_diff}. ObjectClear maintains minimal pixel differences in non-target background areas, which clearly validates its ability to preserve the original background while accurately removing the object and its effects.

\noindent \textbf{Efficiency Comparison with Object Removal Methods.}
ObjectClear is significantly more efficient than both OmniEraser~\citep{wei2025omnieraser} and OmniPaint~\citep{yu2025omnipaint}. 
While the latter two are built on Flux (\(\sim12\)B parameters), ObjectClear is based on SDXL (\(\sim3.7\)B), making our model considerably lighter. 
As a result, ObjectClear runs over \(5\times\) faster during inference, \ie, 1.63 s/image (ObjectClear) vs. 9.42 s/image (OmniPaint) at \(512\times512\) resolution on an A100 GPU.

\subsection{Limitations}
\label{sec:limitations}
\begin{figure*}[h]
	\centering
	\includegraphics[width=0.92\linewidth]{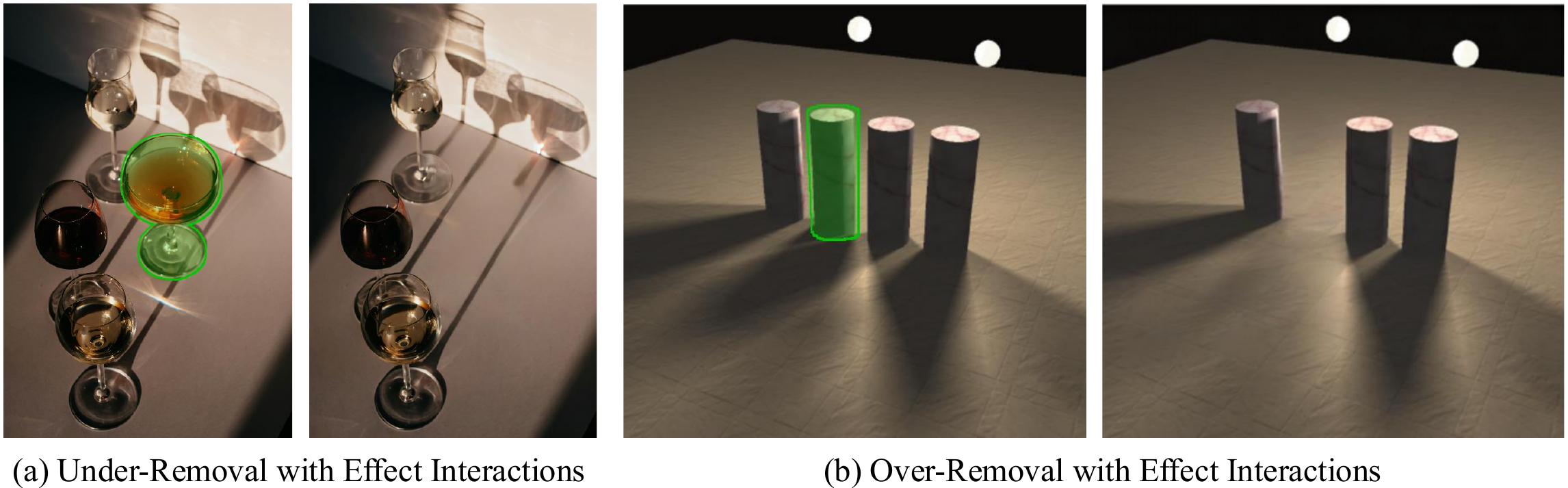}
	\caption{\textbf{Limitations.} 
		In complex scenes where multiple objects may produce overlapping or intertwined effects (\eg, shadows and reflections), our method can consistently remove the objects but may sometimes fail to precisely eliminate the associated effects. This results in either (a) \textit{under-removal of the target effect} or (b) \textit{over-removal of effects belonging to nearby objects}.
	}
	\label{fig:hard_cases}
\end{figure*}

While ObjectClear exhibits strong performance in removing objects and their associated effects, it still faces challenges in highly complex scenarios. Specifically, in cases with overlapping shadows from multiple objects or complex lighting conditions, 
it can be difficult to disentangle which shadows belong to which objects. As a result, the model may fail to remove the shadows of the target object (Fig.~\ref{fig:hard_cases}a) or remove shadows of other objects (Fig.~\ref{fig:hard_cases}b). Effectively disentangling object-specific shadows in such complex scenes remains an important direction for future work.

\begin{figure*}[t]
	\centering
	\includegraphics[width=0.92\linewidth]{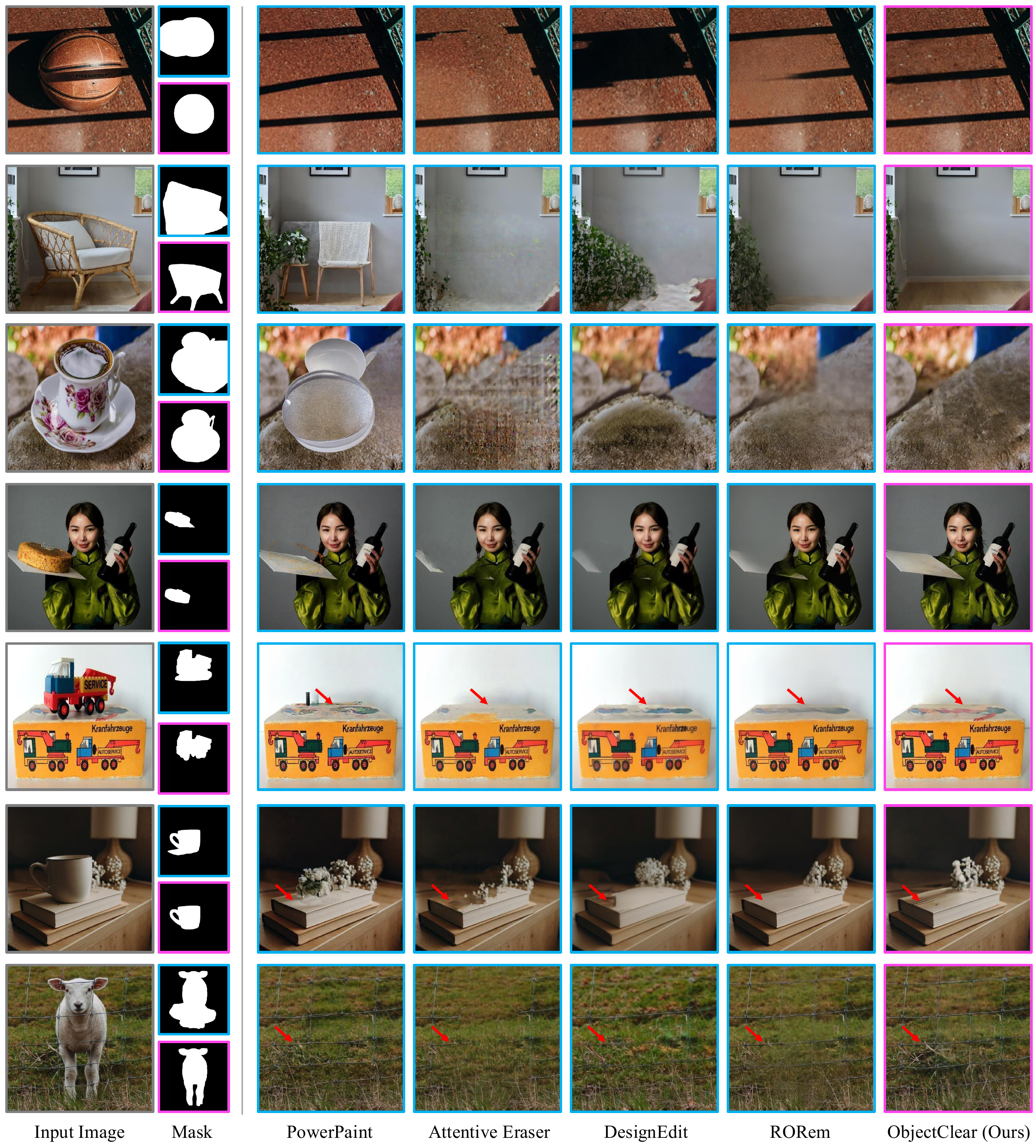}
	\vspace{-2mm}
	\caption{\textbf{Comparison with Image Inpainting Models.} Since some methods struggle to remove object effects when provided with only the object mask, we supply them with our annotated {\textcolor{mycolor_blue}{Object-Effect Mask}} for a fair comparison, while our ObjectClear uses only {\textcolor{mycolor_purple}{Object Mask}} as input. Although these methods are able to remove shadows with the additional effect region, they often introduce undesirable changes to the original background. In contrast, ObjectClear effectively removes the object and its associated effects using only the object mask, while preserving the background content with high fidelity.}
	\label{fig:supp_image_inpainting}
\end{figure*}

\begin{figure*}[t]
	\centering
	\includegraphics[width=0.66\linewidth]{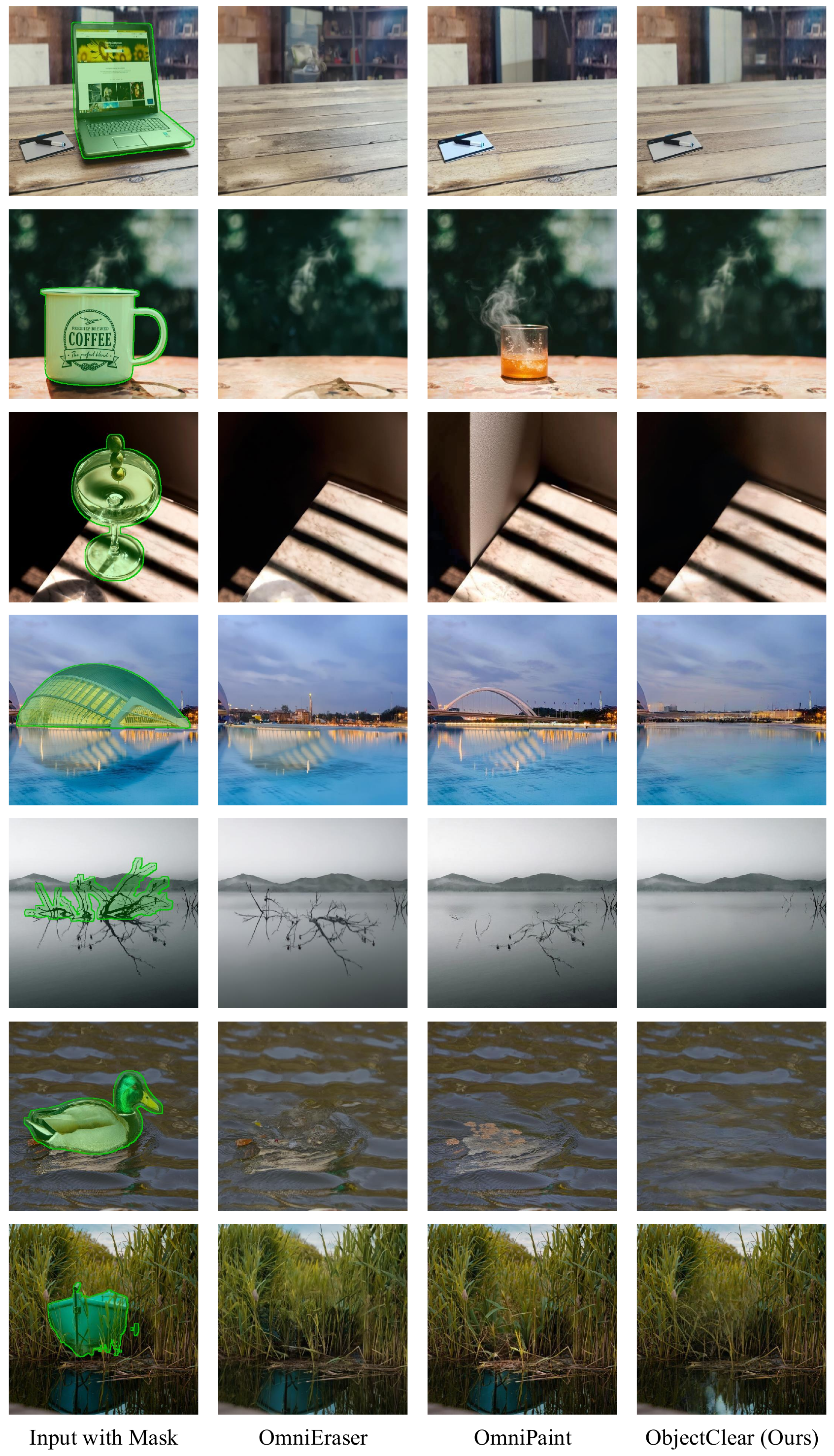}
	\vspace{-2mm}
	\caption{\textbf{Comparison with Object Removal Models.} {All compared methods use Object Masks as input for fair comparison. Our ObjectClear demonstrates superior stability: it not only successfully eliminates the target object and its associated effects, but also avoids unintended changes to the original background, preserving background fidelity with high precision.}}
	\label{fig:supp_object_removal}
\end{figure*}

\begin{figure*}[t]
	\centering
	\includegraphics[width=0.928\linewidth]{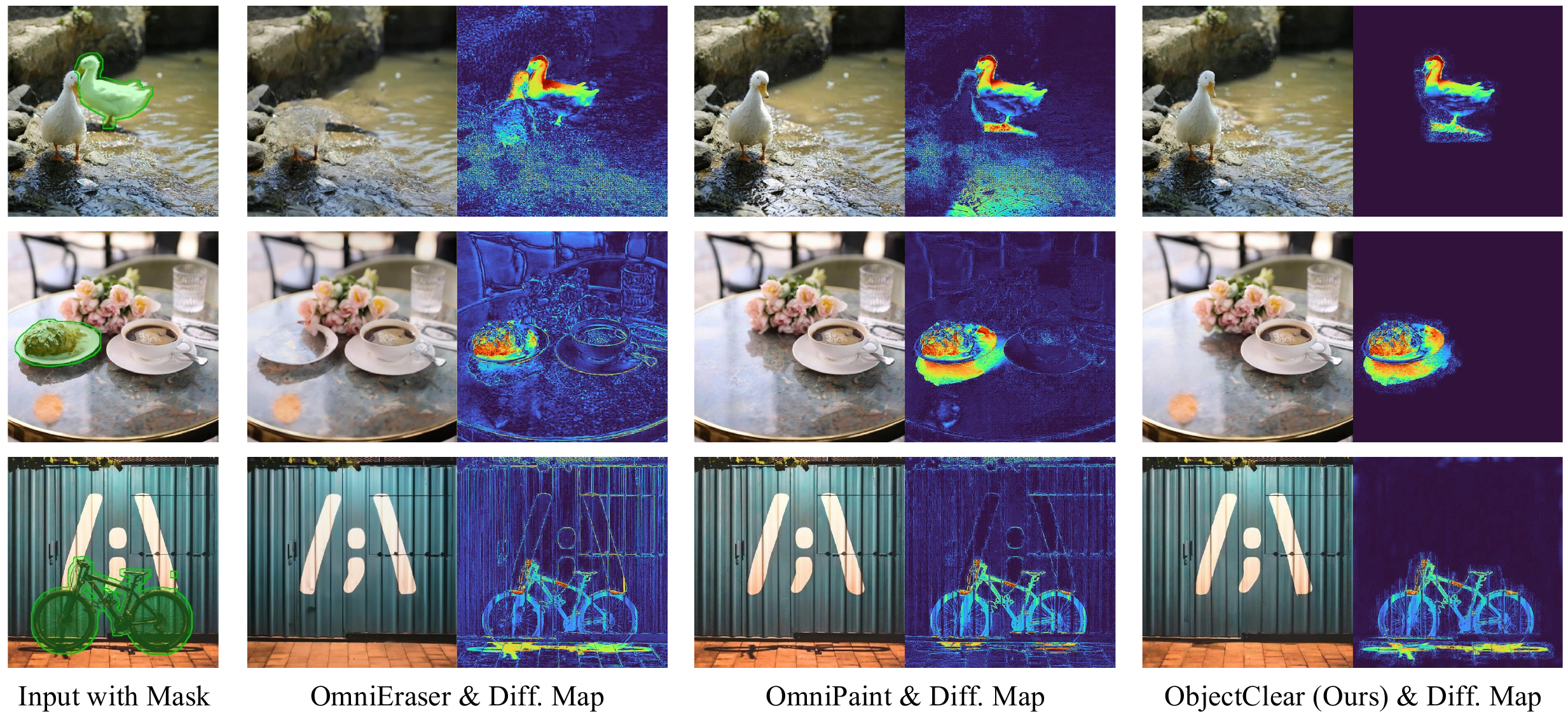}
	\vspace{-2mm}
	\caption{\textbf{Comparison of Object Removal Models via Difference Maps.} We present results of representative object removal methods, and their pixel-wise difference maps relative to the input. ObjectClear successfully removes the target object and its associated shadows while preserving the background. This advantage is clearly reflected by the low pixel-wise differences in non-target areas.}
	\label{fig:supp_object_removal_diff}
\end{figure*}

\end{document}